\documentclass{article}
\pdfoutput=1

\usepackage[numbers, compress]{natbib}
\usepackage[preprint]{neurips_2022}

\usepackage[utf8]{inputenc} 
\usepackage[T1]{fontenc}    
\usepackage{hyperref}       
\usepackage{url}            
\usepackage{booktabs}       
\usepackage{amsfonts}       
\usepackage{nicefrac}       
\usepackage{microtype}      

\usepackage[dvipsnames]{xcolor}
\usepackage{graphicx}
\usepackage{amsmath}
\usepackage{amssymb}
\usepackage{makecell}
\usepackage{multirow}
\usepackage{multicol}
\usepackage{comment}

\usepackage{pifont}

\title{Revealing the Dark Secrets of Masked Image Modeling}

\author{%
  Zhenda Xie\thanks{~Equal Contribution. The work is done when Zhenda Xie, Zigang Geng, and Jingcheng Hu are long-term interns at Microsoft Research Asia. $^{\dag}$ Contact person. }\hspace{1.2mm}$^{13}$, Zigang Geng$^{*23}$, Jingcheng Hu$^{13}$, Zheng Zhang$^{3}$, Han Hu$^{3}$,  Yue Cao$^{3\dag}$ \\
  $^1$Tsinghua University\\
  $^2$University of Science and Technology of China\\
  $^3$Microsoft Research Asia\\
}

\begin{document}

\maketitle

\begin{abstract}
  Masked image modeling (MIM) as pre-training is shown to be effective for numerous vision downstream tasks, but how and where MIM works remain unclear. In this paper, we compare MIM with the long-dominant supervised pre-trained models from two perspectives, the visualizations and the experiments, to uncover their key representational differences. From the visualizations, we find that MIM brings locality inductive bias to all layers of the trained models, but supervised models tend to focus locally at lower layers but more globally at higher layers. That may be the reason why MIM helps Vision Transformers that have a very large receptive field to optimize. Using MIM, the model can maintain a large diversity on attention heads in all layers. But for supervised models, the diversity on attention heads almost disappears from the last three layers and less diversity harms the fine-tuning performance. From the experiments, we find that MIM models can perform significantly better on geometric and motion tasks with weak semantics or fine-grained classification tasks, than their supervised counterparts. Without bells and whistles, a standard MIM pre-trained SwinV2-L could achieve state-of-the-art performance on pose estimation (78.9 AP on COCO test-dev and 78.0 AP on CrowdPose), depth estimation (0.287 RMSE on NYUv2 and 1.966 RMSE on KITTI), and video object tracking (70.7 SUC on LaSOT). For the semantic understanding datasets where the categories are sufficiently covered by the supervised pre-training, MIM models can still achieve highly competitive transfer performance. With a deeper understanding of MIM, we hope that our work can inspire new and solid research in this direction.
\end{abstract}

\section{Introduction}
Pre-training of effective and general representations applicable to a wide range of tasks in a domain is the key to the success of deep learning. 
In computer vision, supervised classification on ImageNet~\cite{deng2009imagenet} has long been the dominant pre-training task which is manifested to be effective on a wide range of vision tasks, especially on the semantic understanding tasks, such as image classification~\cite{donahue2014decaf,kornblith2019better,alex2019big,dosovitskiy2020vit,liu2021swin}, object detection~\cite{sermanet2013overfeat,girshick2014rich,ren2015faster,Mask-rcnn}, semantic segmentation~\cite{long2015fully,wang2018non}, video action recognition~\cite{simonyan2014two,tran2015learning,carreira2017quo,liu2021video} and so on.
Over the past several years, “masked signal modeling”, which masks a portion of input signals and tries to predict these masked signals, serves as a universal and effective self-supervised pre-training task for various domains, including language, vision, and speech.
After (masked) language modeling repainted the NLP field~\cite{devlin2018bert,liu2019roberta}, recently, such task has also been shown to be a competitive challenger to the supervised pre-training in computer vision~\cite{chen2020imagegpt,dosovitskiy2020vit,bao2021beit,he2021masked,zhou2021ibot,xie2021simmim}. That is, masked image modeling (MIM) pre-trained models achieve very high fine-tuning accuracy on a wide range of vision tasks of different nature and complexity.

However, there still remain several questions:
\begin{enumerate}
    \item What are the key mechanisms that contribute to the excellent performance of MIM?
    \item How transferable are MIM and supervised models across different types of tasks, such as semantic understanding, geometric and motion tasks?
\end{enumerate}
To investigate these questions, we compare MIM with supervised models from two perspectives, the visualization perspective and the experimental perspective, trying to uncover key representational differences between these two pre-training tasks and deeper understand the behaviors of MIM pre-training.

We start with studying the attention maps of the pre-trained models.
Firstly, we visualize the averaged attention distance in MIM models, and we find that \textbf{masked image modeling brings locality inductive bias to the trained model, that the models tend to aggregate near pixels in part of the attention heads,} and the locality strength is highly correlated with the masking ratio and masked patch size in the pre-training stage. But the supervised models tend to focus locally at lower layers but more globally at higher layers.

We next probe how differently the attention heads in MIM trained Transformer behave. We find that \textbf{different attention heads tend to aggregate different tokens on all layers in MIM models}, according to the large KL-divergence on attention maps of different heads.
But for supervised models, the diversity on attention heads diminishes as the layer goes deeper and almost disappears in the last three layers. We drop the last several layers for supervised pre-trained models during fine-tuning and find that it benefits the fine-tuning performance on downstream tasks, however this phenomenon is not observed for MIM models. That is, \textbf{less diversity on attention heads would somewhat harm the performance on downstream tasks}. 

Then we examine the representation structures in the deep networks of MIM and supervised models via the similarity metric of Centered Kernel Alignment (CKA)~\cite{cka}. We surprisingly find that \textbf{in MIM models, the feature representations of different layers are of high similarity, that their CKA values are all very large (e.g., [0.9, 1.0]).} But for supervised models, as in~\cite{raghu2021vision}, different layers learn different representation structures, that their CKA similarities vary greatly (e.g., [0.5,1.0]). To further verify this, we load the pre-trained weights of randomly shuffled layers during fine-tuning and find that supervised pre-trained models suffer more than the MIM models.

From the experimental perspective, a fundamental pretraining task should be able to benefit a wide range of tasks, or at least it is important to know for which types of tasks MIM models work better than the supervised counterparts. To this end, we conduct a large-scale study by comparing the fine-tuning performance of MIM and supervised pre-trained models, on three types of tasks, semantic understanding tasks, geometric and motion tasks, and the combined tasks which simultaneously perform both.

For semantic understanding tasks, we select several representative and diverse image classification benchmarks, including Concept Generalization (CoG) benchmark~\cite{sariyildiz2021cog}, the widely-used 12-dataset benchmark~\cite{kornblith2019better}, as well as a fine-grained classification dataset iNaturalist-18~\cite{van2018inaturalist}. 
For the classification datasets whose categories are sufficiently covered by ImageNet categories (e.g. CIFAR-10/100), supervised models can achieve better performance than MIM models. However, for other datasets, such as fine-grained classification datasets (e.g., Food, Birdsnap, iNaturalist), or datasets with different output categories (e.g., CoG), most of the representation power in supervised models is difficult to transfer, thus MIM models remarkably outperform supervised counterparts.

For geometric and motion tasks that require weaker semantics and high-resolution object localization capabilities, such as pose estimation on COCO~\cite{lin2014microsoft} and CrowdPose~\cite{li2019crowd}, depth estimation on NYUv2~\cite{nathan2012nyuv2} and KITTI~\cite{andreas2013kitti}, and video object tracking on GOT10k~\cite{huang2021got10k}, TrackingNet~\cite{muller2018tracknet}, and LaSOT~\cite{fan2019lasot}, MIM models outperform supervised counterparts by large margins. Note that, without bells and whistles, Swin-L with MIM pre-training could achieve state-of-the-art performance on these benchmarks, e.g., $80.5$ AP on COCO $val$, $78.9$ AP on COCO $test$-$dev$, and $78.0$ AP on CrowdPose of pose estimation, $0.287$ RMSE on NYUv2 and $1.966$ RMSE on KITTI of depth estimation, and $70.7$ SUC on LaSOT of video object tracking.

We select object detection on COCO as the combined task which simultaneously performs both semantic understanding and geometric learning. For object detection on COCO, MIM models would outperform supervised counterparts. 
Via investigating the training losses of object classification and localization, we find that MIM models help localization task converge faster, and supervised models benefit more for object classification, that categories of COCO are fully covered by ImageNet.

In general, MIM models can perform significantly better on geometric/motion tasks with weak semantics or fine-grained classification tasks, than the supervised counterparts. For tasks/datasets where supervised models are good at transfer, MIM models can still achieve highly competitive transfer performance. It seems time to embrace masked image modeling as a general-purpose pre-trained model. We hope our paper can drive this belief deeper in the community and inspire new and solid research in this direction.

\section{Background}

\noindent\textbf{Masked Image Modeling.}
Masked image modeling (MIM) is a sub-task of masked signal prediction, that masks a portion of input images, and lets the deep networks predict the masked signals conditioned on the visible ones.
We use SimMIM~\cite{xie2021simmim}, a simple framework for masked image modeling, as the exampled framework of pre-trained image models in our visualizations and experiments, because it is simple, effective, and generally applicable. 
SimMIM consists of four major components with simple designs: 1) random masking with a moderately large masked patch size (e.g., 32); 2) the masked tokens and image tokens are fed together to the encoder; 3) the prediction head is as light as a linear layer: 4) directly predicting raw pixels of RGB values as the target with the $\ell_1$ loss of direct regression. 
With these simple designs, SimMIM can achieve state-of-the-art performance on ImageNet-1K classification, COCO object detection, and ADE-20K semantic segmentation.
Note that, the SimMIM framework could be directly applied to different types of backbone architectures, such as Vision Transformer (ViT)~\cite{dosovitskiy2020vit}, Swin Transformer~\cite{liu2021swin}, and ConvNets~\cite{he2016resnet,replknet}. This property enables us to study the characteristics of MIM under different types of backbone architectures, as well as in multiple types of downstream tasks.

\noindent\textbf{Backbone Architectures.} Masked image modeling is mostly studied in the Transformer architectures, thus the major understandings and experiments in this paper are performed on Vision Transformers (ViT)~\cite{dosovitskiy2020vit} and Swin Transformers~\cite{liu2021swin,swinv2}. Due to the simple and clear architecture designs of ViT, most of the visualizations are performed on ViT, shown in Section~\ref{sec:vis}. Due to the general-purpose property of Swin Transformer, most of the experiments on different downstream tasks are conducted on Swin Transformer, shown in Section~\ref{sec:exp}.

\section{Visualizations}\label{sec:vis}

\subsection{Revealing the Properties of Attention Maps}

Attention mechanism~\cite{bahdanau2014neural} has been an exceptional component in deep networks. It is naturally interpretable since attention weights have a clear meaning: how much each token is weighted when determining the output representation of the current token. Fortunately, most MIM pre-trained models~\cite{dosovitskiy2020vit,bao2021beit,he2021masked,zhou2021ibot,xie2021simmim} are established upon the Vision Transformers, where self-attention block is its major component. Here we start with studying the attention maps of the pre-trained models from three angles: (a) averaged attention distance to measure whether it is local attention or global attention; (b) entropy of attention distribution to measure whether it is focused attention or broad attention; (c) KL divergence of different attention heads to investigate that attention heads are attending different tokens or similar ones.

\subsubsection{Local Attention or Global Attention?}
Images are observed to exhibit strong locality: pixels near each other tend to be highly correlated~\cite{hubel1962receptive}, motivating the use of local priors in a wide range of visual perception architectures~\cite{fukushima1975cognitron,lecun1999object,alexnet,he2016resnet,liu2021swin}. In the era of Vision Transformers, the usefulness of local priors has still undergone rich discussions and trials~\cite{dosovitskiy2020vit,liu2021swin,li2021localvit}. Thus it is valuable to investigate whether MIM models bring the locality inductive bias to the models. We do this by computing averaged attention distance in each attention head of each layer. 

\begin{figure}
    \centering
    \includegraphics[width=1.\textwidth]{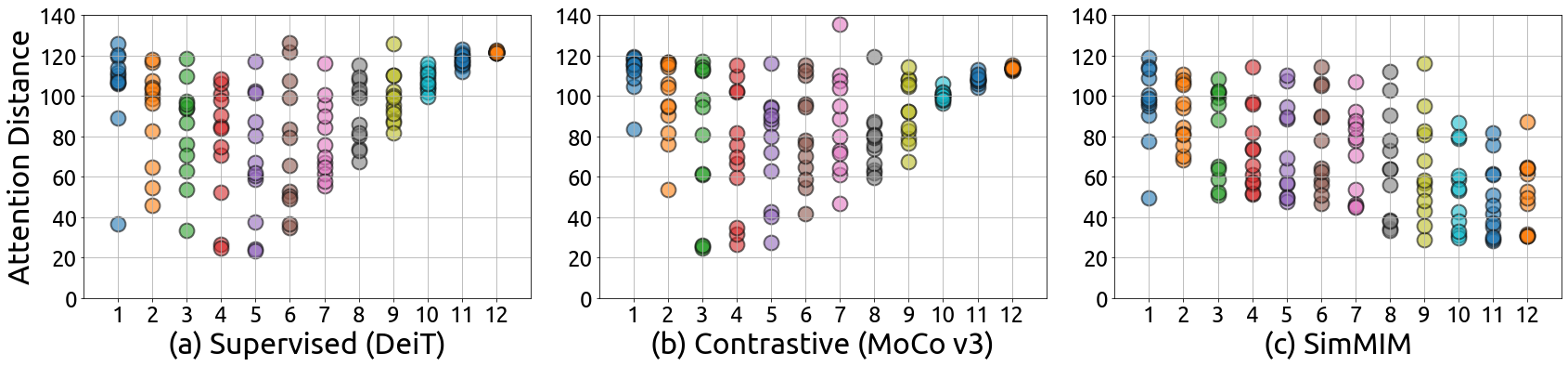}
	    \vspace{-1.5em}
    \caption{The averaged attention distance in different attention heads (dots) w.r.t the layer number on supervised model (a), contrastive learning model (b), and SimMIM model (c) with ViT-B as the backbone architecture.}
    \label{fig:locality}
\end{figure}

\begin{figure}
    \centering
    \includegraphics[width=1.\linewidth]{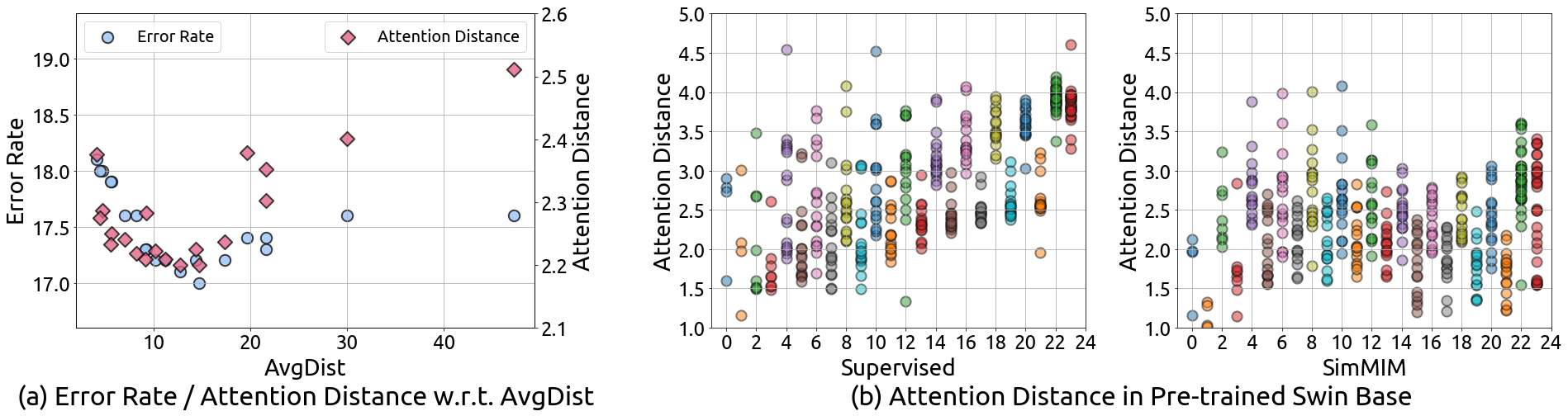}
	    \vspace{-1.5em}
    \caption{(a) The error rate of fine-tuning on ImageNet-1K (blue circle \textcolor{blue}{$\circ$}) and averaged attention distance (red diamond \textcolor{red}{$\diamond$}) w.r.t AvgDist (averaged distance of masked pixels to the nearest visible pixels) with Swin-B as the backbone. Points (\textcolor{red}{$\diamond$} or \textcolor{blue}{$\circ$}) denote the SimMIM models with different masking ratios and masked patch sizes. (b) The averaged attention distance in different attention heads (dots) w.r.t the layer number on supervised model (b1) and SimMIM model (b2) with Swin-B as the backbone.}
    \label{fig:avgdist}
\end{figure}

Results of the averaged attention distance in different attention heads (dots) w.r.t the layer number, on supervised model (DeiT), contrastive learning model (MoCo v3) and SimMIM model with ViT-B as backbone are shown in Figure~\ref{fig:locality}. 
We find that the supervised model tends to focus locally at lower layers but more globally at higher layers, which well matches the observations in ViT~\cite{dosovitskiy2020vit}. Surprisingly, the contrastive learning model acts very similarly to the supervised counterpart. This may also be understandable, since MoCo v3 has a very high linear evaluation accuracy on ImageNet-1K (76.7\% of top-1 accuracy), which indicates that the features of the last layer of MoCo v3 are very similar to that of the supervised counterpart.
But for the model trained by SimMIM, its behavior is significantly different to supervised and contrastive learning models. Each layer has diverse attention heads that tend to aggregate both local and global pixels, and the average attention distance is similar to the lower layers of the supervised model. As the number of layers gets deeper, the averaged attention distance becomes even slightly smaller. That is, MIM brings locality inductive bias to the trained model, that the models tend to aggregate near pixels in part of the attention heads. Also, a similar observation could be observed with Swin-B as the backbone, as shown in Figure~\ref{fig:avgdist}(b).

SimMIM~\cite{xie2021simmim} designed a new metric, AvgDist, which measures the averaged Euclidean distance of masked pixels to the nearest visible ones and indicates the task difficulty and effectiveness of MIM depending on the masking ratio and masked patch size. As shown in Figure~\ref{fig:avgdist}(a), AvgDist is a good indicator that the entries of high fine-tuning accuracy roughly distribute in a range of [10, 20] of AvgDist, while entries with smaller or higher AvgDist perform worse. Interestingly, in the range of [10, 20] of AvgDist, we can also observe a small averaged attention distance. That is, a moderate prediction distance in MIM will bring a greater strength of locality and incur a better fine-tuning performance.

\subsubsection{Focused Attention or Broad Attention?}

\begin{figure}[t]
    \centering
    \includegraphics[width=1.\linewidth]{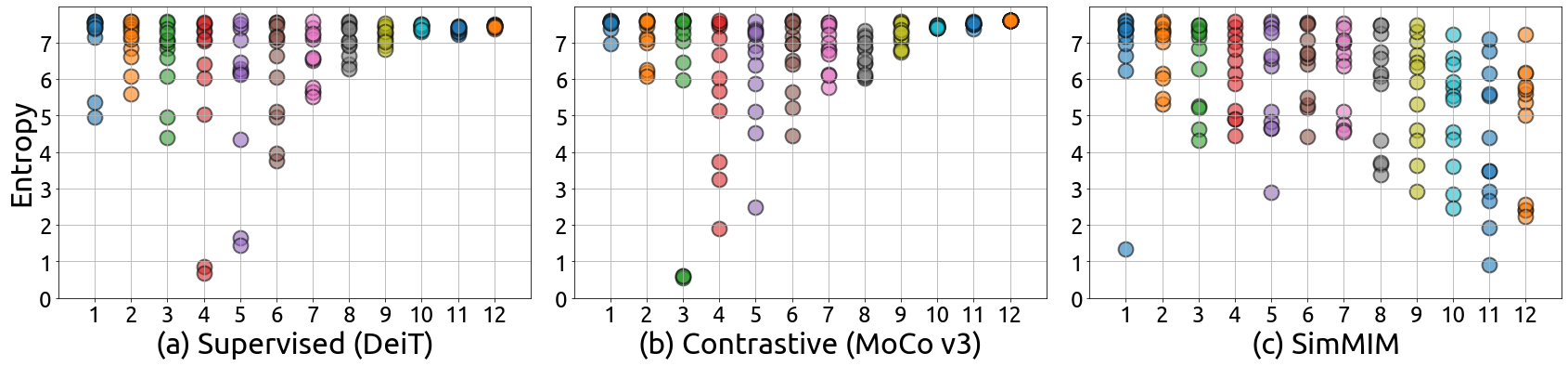}
	    \vspace{-1.5em}
    \caption{The entropy of each head's attention distribution w.r.t the layer number on (a) supervised model, (b) contrastive learning model,  and (c) SimMIM model with ViT-B as the backbone.}
    \label{fig:entropy}
\end{figure}

We then measure the attention maps on whether attention heads focus on a few tokens or attend broadly over many tokens, via averaging the entropy of each head's attention distribution. Results of entropy values w.r.t different layers of three pre-trained models, supervised model (DeiT), contrastive learning model (MoCo v3), and MIM model (SimMIM) with ViT-B as the backbone, are shown in Figure~\ref{fig:entropy}. For supervised models, we find that some attention heads in lower layers have very focused attention, but in higher layers, most attention heads focus very broadly. The contrastive model still behaves very similarly to the supervised model. But for the MIM model, the entropy values in different attention heads are diverse in all layers, that some attention heads are more focused and some heads have very broad attention.

\subsubsection{Diversity on Attention Heads}
\label{subsubsection:diversity}

\begin{figure}
    \centering
    \includegraphics[width=1.\linewidth]{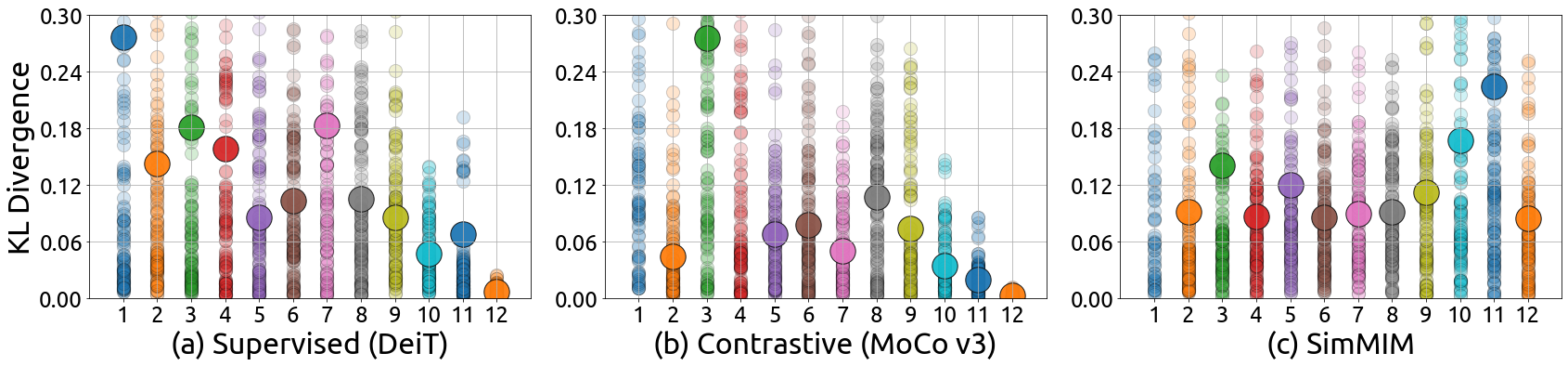}
	    \vspace{-1.5em}
    \caption{The KL divergence between attention distributions of different heads (small dots) and the averaged KL divergence (large dots) in each layer w.r.t the layer number on (a) supervised model, (b) contrastive learning model, and (c) SimMIM model with ViT-B as the backbone architecture.}
    \label{fig:kl-divergence}
\end{figure}

From the previous two sub-sections, we observe a similar phenomenon, that is, for the supervised model, the attention distance or entropy of attention heads in the last few layers seem to be similar, while for the MIM model, different heads in all layers behave more diversely. Therefore, we want to further explore whether the different heads pay attention to different/similar tokens, via computing the 
Kullback–Leibler (KL) divergence~\cite{kldivergence} between the attention maps of different heads in each layer. 

Results of KL divergence between attention distributions of different heads w.r.t different layers of three pre-trained models, supervised model (DeiT), contrastive learning model (MoCo v3), and MIM model (SimMIM) with ViT-B as the backbone, are shown in Figure~\ref{fig:kl-divergence}. As we expect, different attention heads tend to aggregate different tokens on all layers in MIM models, according to the large KL-divergence on attention maps of different heads. But for supervised models and contrastive learning models, the diversity on attention heads becomes smaller as the layer goes deeper and almost disappears from the last three layers.

\begin{figure*}
\centering
    \includegraphics[width=1.\textwidth]{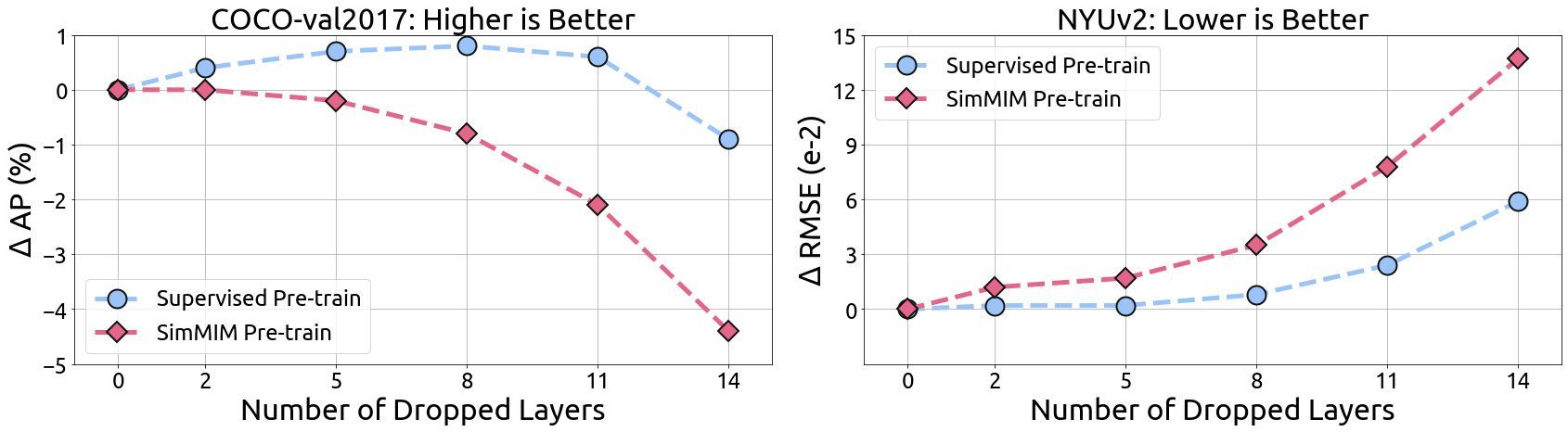}
	    \vspace{-1.5em}
   \caption{The performance of the COCO $val2017$ pose estimation (left) and NYUv2 depth estimation (right) when we drop several last layers of the SwinV2-B backbone. When the model becomes smaller, the performance of the supervised pre-trained model increases on the pose estimation and keeps the same on the depth estimation. The last layers in the supervised pre-trained model lose diversity across different attention heads and are harmful to the downstream tasks.
   }
\label{fig:drop-block}
\end{figure*}

Intuitively, losing diversity across different attention heads may limit the capacity of the model. To investigate whether the loss of diversity on attention heads has any adverse effect, we gradually drop layers from the end, and only load previous layers when fine-tuning the model for the downstream tasks of COCO $val2017$ pose estimation and NYUv2 depth estimation. From Figure~\ref{fig:drop-block}, we can observe that when we drop two to eight layers, although the model becomes smaller, the performance of the supervised pre-trained model on COCO $val2017$ pose estimation is better than the baseline, and the performance on NYUv2 depth estimation is comparable with the baseline. This shows that in the supervised pre-trained model, the last layers with small diversity on attention heads indeed affect the performance of downstream tasks. The detailed setup of this experiment is in the Appendix.

\subsection{Investigating the Representation Structures via CKA similarity}
\begin{figure}
    \centering
    \includegraphics[width=1.\linewidth]{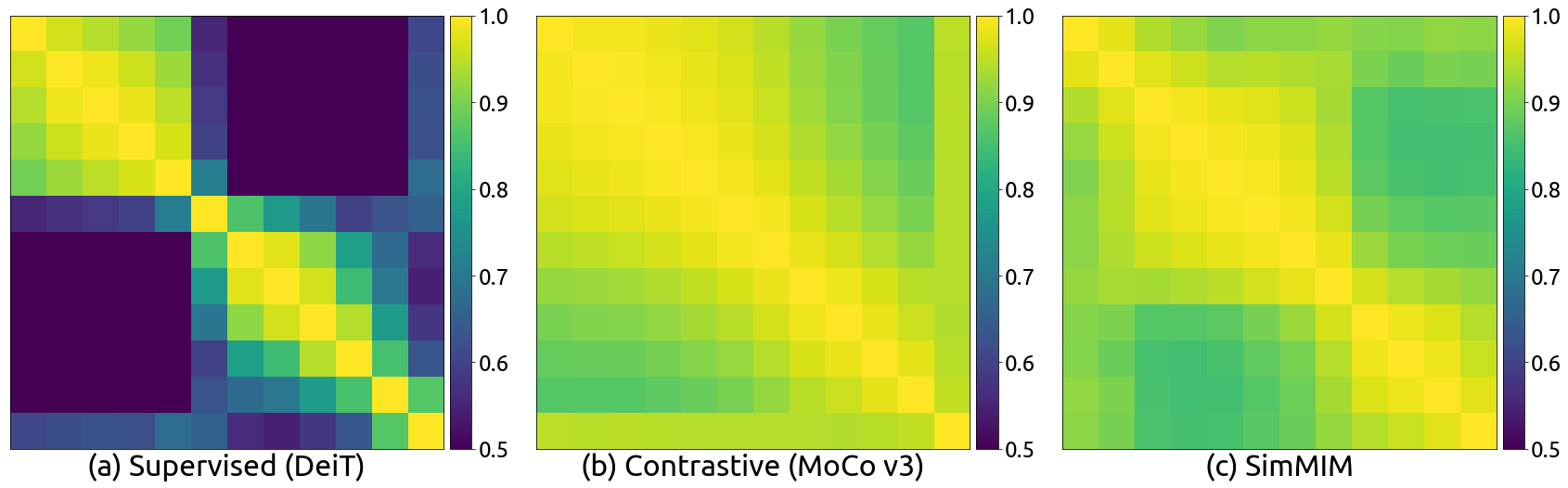}
	    \vspace{-1.5em}
    \caption{The CKA heatmap between the feature maps of different layers of (a) supervised model, (b) contrastive learning model, and (c) SimMIM model with ViT-B as the backbone architecture.}
    \label{fig:cka_layer}
\end{figure}

Studying the behaviors of attention mechanisms is analyzing inside the block, from a micro perspective. Next, we hope to study from a macro perspective of deep networks, such as studying the similarity between feature maps across different layers via the CKA similarity~\cite{cka}. 
Results of CKA similarity between feature representations of different layers of three pre-trained models, supervised model (DeiT), contrastive learning model (MoCo v3), and MIM model (SimMIM) with ViT-B as the backbone, are shown in Figure~\ref{fig:cka_layer}. We surprisingly find that in MIM models, the representation structures of different layers are almost the same, that their CKA similarities are all very large (e.g., [0.9, 1.0]). But for supervised models, as in~\cite{raghu2021vision}, different layers learn different representation structures, that their CKA similarities vary greatly (e.g., [0.5,1.0]). Different from previous visualizations, MoCo v3 behaves similarly to SimMIM in this case.

\label{subsubsection:similar-layers}

To further verify this observation, we load the pre-trained weights of randomly shuffled layers and fine-tune the model for the downstream tasks of COCO pose estimation and NYUv2 depth estimation. 
We observe that by loading the models with the randomly sampled layers, the performance on 1K-MIM drops from 75.5 to 75.2 (-0.3) on pose estimation and 0.382 to 0.434 (-0.052) on depth estimation. But supervised pre-trained models suffer more than the MIM models, which drops from 75.8 to 74.9 (-0.9) on pose estimation, and 0.376 to 0.443 (-0.067) on depth estimation. 
The detailed setup of this experiment is in the Appendix.

\section{Experimental Analysis on Three Types of Downstream Tasks}\label{sec:exp}
\label{section:experiment}

In this section, we conduct a large-scale study by comparing the fine-tuning performance of MIM and supervised pre-trained models, on three types of tasks, semantic understanding tasks (e.g., image classification in different domains), geometric and motion tasks (e.g., pose/depth estimation, and video object tracking), and the combined tasks which simultaneously perform both types of tasks (e.g., object detection).
We use 8 NVIDIA V100 GPUs for our experiments.

\subsection{Semantic Understanding Tasks}

\renewcommand{\arraystretch}{1.15}
\begin{table}[t]
  \centering
  \addtolength{\tabcolsep}{-2.3pt}
  \begin{tabular}{c|ccccc|m{0.8cm}<{\centering}m{1.0cm}<{\centering}m{0.8cm}<{\centering}m{1.0cm}<{\centering}m{1.7cm}<{\centering}|m{0.9cm}<{\centering}}
    \bottomrule
    \multirow{2}{*}{pre-train}  & \multicolumn{5}{c|}{Concept Generalization (CoG)} & 
    \multicolumn{5}{c|}{Kornlith et al's 12 datasets (K12)}  & 
    \multirow{2}{*}{iNat18}\\
    \cline{2-11}
   & $L_1$ & $L_2$ & $L_3$ & $L_4$ & $L_5$ & Food & Birdsnap & Cars & Aircraft & {Average ($7$)} & \\
    \hline
    1K-SUP & $79.4$ &  $76.2$ & $72.7$ & $72.5$ & $68.4$ & $93.2$ & $81.8$ & $88.6$ & $83.0$ & $89.7$ & $77.7$ \\
    1K-MIM & $79.6$ & $77.1$ & $73.6$ & $73.0$ & $69.1$ & $94.2$ & $83.7 $ & $89.2$ & $83.5$ & $86.1$ & $79.6$ \\
    \toprule
  \end{tabular}
  \caption{Comparisons of MIM and supervised (SUP) pre-trained models on semantic understanding tasks with SwinV2-B as the backbone. We follow~\cite{kornblith2019better} to report top-1 accuracy ($\uparrow$) and mean per-class accuracy ($\uparrow$) for specific datasets. Results on the multi-label dataset Pascal Voc 2007 are not included, whose evaluation metric is not compatible with others.
  }
  \label{tab-overall-zeroshot}
  \vspace{-1.0em}
\end{table}

For semantic understanding tasks, we select several representative and diverse image classification benchmarks, including Concept Generalization (CoG) benchmark~\cite{sariyildiz2021cog}, the widely-used 12-dataset benchmark~\cite{kornblith2019better}, as well as a fine-grained classification dataset iNaturalist-18~\cite{van2018inaturalist}. 

\noindent\textbf{Setup.} The CoG benchmark consists of five 1k-category datasets split from ImageNet-21K, which has an increasing semantic gap with ImageNet-1K, from $L_1$ to $L_5$.
On the CoG dataset, we search for the best hyper-parameters based on the top-1 accuracy of the  $L_1$ validation set and then apply the best setting to CoG $L_2$ to $L_5$ to report the top-1 accuracy.  
On the K12 dataset, we adopt standard splits of train/val/test sets as in~\cite{kornblith2019better}. We use the training set to fine-tune the models, use the validation set to search for the best hyper-parameters, and then train the models on the merged training and validation sets using the best setting. Following~\cite{kornblith2019better}, we report mean-per-class accuracy for Aircraft, Pets, Caltech-101, Oxford 102 Flowers and top-1 accuracy for other datasets. 
The iNat18 dataset includes 437,513 training images and 24,426 validation images, with more than 8,000 categories. We fine-tune the pre-trained models using the training set and report the top-1 accuracy on the validation set. 
For all datasets, we choose learning rate, weight decay, layer decay, and DropPath~\cite{huang2016deep} on the valid set respectively for the MIM pre-trained model and the supervised pre-trained model. We use the AdamW optimizer~\cite{Loshchilov2019adamw} and cosine learning rate schedule. We train the model for $100$ epochs with $20$ warm-up epochs. The input image size is $224 \times 224$. Other detailed setups of these datasets are in the Appendix.

\noindent\textbf{Results.} Results of different semantic understanding tasks are shown in Table~\ref{tab-overall-zeroshot}.
For the classification datasets whose categories are sufficiently covered by ImageNet categories (e.g. CIFAR-10/100), supervised models can achieve better performance than MIM models as pre-training. However, for other datasets, such as fine-grained classification datasets (e.g., Food, Birdsnap, iNaturalist), or datasets with different output categories (e.g., CoG), most of the representation power in supervised models is difficult to transfer; thus MIM models remarkably outperform supervised counterparts.

\subsection{Geometric and Motion Tasks}
\label{subsection:exp-details}

    We study how MIM models perform on the geometric and motion tasks that require the ability to localize the objects and are less dependent on semantic information. We select several benchmarks, such as pose estimation on COCO~\cite{lin2014microsoft} and CrowdPose~\cite{li2019crowd},  depth estimation on NYUv2~\cite{nathan2012nyuv2} and KITTI~\cite{andreas2013kitti}, and video object tracking on GOT10k~\cite{huang2021got10k}, TrackingNet~\cite{muller2018tracknet}, and LaSOT~\cite{fan2019lasot}.

\noindent\textbf{Setup.}
	For pose estimation on COCO and Crowdpose, we use the standard splits for training and evaluation and report the AP based on OKS as the evaluation metric. We use the standard person detection results from ~\cite{xiao2018sbaseline}.
    We follow Simple Baseline~\cite{xiao2018sbaseline}, which upsamples the last feature of the backbone by deconvolutions and predicts the heatmaps at 4$\times$ resolution. 
	The data augmentations include random flipping, half body transformation, random scale, random rotation, grid dropout, and color jittering. The input image size is $256\times 256$ by default. We use the AdamW~\cite{Loshchilov2019adamw} optimizer with the base learning rate $5e$-$4$ and the weight decay $5e$-$2$. The learning rate is dropped to $5e$-$5$ at the $120th$ epoch. We train the models for $150$ epochs. We use a layer decay of $0.9$/$0.85$ for Swin-B/L and the DropPath~\cite{huang2016deep} of $0.3$/$0.5$ for Swin-B/L.

	For depth estimation on NYUv2 and KITTI, we use the standard splits and report the RMSE (Root Mean Square Error) as the evaluation metric.
	To compare with the previous works~\cite{ranftl2021dpt, kim2022glpdepth}, we set the maximum range as $10$m/$80$m for NYUv2/KITTI.
	The head of the depth estimation is the same as that of the pose estimation and is comprised of deconvolutions. 
     Similar to the GLPDepth~\cite{kim2022glpdepth}, we use the following data augmentations: random horizontal flip, random brightness/gamma/hue/saturation/value and random vertical CutDepth. We randomly crop the images to $480\times 480$ / $352\times 352$ size for NYUv2/KITTI dataset. The optimizer, layer decay, and DropPath is the same as the pose estimation. The learning rate is scheduled via polynomial strategy with a factor of $0.9$ with a minimal value of $3e$-$5$ and a maximum value of $5e$-$4$. The total number of epochs is $25$. We use the flip testing and sliding window test.
	 
	Following the previous methods~\cite{lin2021swintrack,cui2022mixfmr}, we train the models on the train splits of four datasets GOT10k~\cite{huang2021got10k}, TrackingNet~\cite{muller2018tracknet}, LaSOT~\cite{fan2019lasot}, and COCO~\cite{lin2014microsoft} and report the success score (SUC) for the TrackingNet dataset and LaSOT dataset, and the average overlap (AO) for GOT10k. 
	We use the SwinTrack~\cite{lin2021swintrack} to train and evaluate our pre-trained models with the same data augmentations, training, and inference settings. 
	We sample $131072$ pairs per epoch and train the models for $300$ epochs. We use the AdamW optimizer with a learning rate of $5e$-$4$ for the head, a learning rate of $5e$-$5$ for the backbone, and a weight decay of $1e$-$4$. We decrease the learning rate by a ratio of $0.1$ at the $210$th epoch. We set the sizes of search images and templates as $224\times 224$ and $112\times 112$. 

	\renewcommand{\arraystretch}{1.15}
	\begin{table*}[t]
		\centering\setlength{\tabcolsep}{6pt}
		\footnotesize
		\begin{tabular}{m{1cm}<{\centering}m{1.3cm}<{\centering} | m{0.8cm}<{\centering}m{0.8cm}<{\centering}m{0.8cm}<{\centering} | m{1.0cm}<{\centering}m{1.0cm}<{\centering} | m{0.9cm}<{\centering} m{0.9cm}<{\centering}m{0.9cm}<{\centering}}
			\bottomrule
			\multirow{3}*{backbone} & \multirow{3}*{pre-train} & \multicolumn{3}{c|}{Pose Estimation} & \multicolumn{2}{c|}{Depth Estimation} & \multicolumn{3}{c}{Video Object Tracking} \\
			\cline{3-10}
			 & & COCO $val$ & COCO $test$ & Crowd-Pose & NYUv2 & KITTI & GOT10k $test$  & Track-Net & LaSOT\\
			\hline
			\multirow{3}*{SwinV2-B} & $1$K-SUP &$75.2$ & $74.5$ & $70.7$ & $0.352$ & $2.313$ & $70.1$ & $81.5$ & $69.4$\\
			 & $22$K-SUP & $75.9$ & $75.1$ & $72.2$ & $0.335$ & $2.240$ & $69.9$ & $81.0$ & $67.8$\\
			 & $1$K-MIM & $\textbf{77.6}$ & $\textbf{76.7}$ & $\textbf{74.9}$ & $\textbf{0.304}$ & $\textbf{2.050}$ & $\textbf{70.8}$ & $\textbf{82.0}$ & $\textbf{70.0}$\\
			\hline 
			\multirow{2}*{SwinV2-L} & $22$K-SUP & $76.5$ & $75.7$ & $72.7$ & $0.334$ & $2.150$ & $71.1$  & $81.5$& $69.2$\\
			 & $1$K-MIM & $\textbf{78.1}$ & $\textbf{77.2}$ & $\textbf{75.5}$ & $\textbf{0.287}$ & $\textbf{1.966}$ & $\textbf{72.9}$ & $\textbf{82.5}$& $\textbf{70.7}$\\
			 \hline
			 \multicolumn{2}{c|}{\multirow{2}*{Representative methods}} & \multicolumn{3}{c|}{HRFormer~\cite{yuan2021hrformer}} & \multicolumn{2}{c|}{BinsFormer~\cite{li2022binsdepth}} & \multicolumn{3}{c}{MixFormer~\cite{cui2022mixfmr}}\\
			 \cline{3-10}
			 & & $77.2$ & $76.2$ & $72.5$ & $0.330$ & $2.098$ & $75.6$ & $83.9$ & $70.1$\\
			\toprule
		\end{tabular}
		\caption{Comparisons of MIM and supervised (SUP) pre-trained models on the geometric and motion tasks. We report the AP ($\uparrow$) for the pose estimation tasks, RMSE ($\downarrow$) for the monocular depth estimation tasks, AO ($\uparrow$) for the GOT10K dataset, and SUC ($\uparrow$) for the TrackingNet dataset and LaSOT tracking dataset. The best results among the different pre-trained models are shown in the \textbf{bold} text. We provide the best results of the representative methods for reference.} 
		\label{tab:pose-depth-track}
		\vspace{-0.8em}
	\end{table*}
	
	\noindent\textbf{Results.}
	From Table ~\ref{tab:pose-depth-track}, for the pose estimation, MIM models pre-trained with ImageNet-1K surpass supervised counterparts by large margins, $2.4$ AP on COCO $val$, $2.2$ AP on COCO $test$-$dev$, and $4.2$ AP on CrowdPose dataset which contains more crowded scenes. Even if the supervised models are pre-trained with ImageNet-22K, the performances are still worse than MIM models pre-trained with ImageNet-1K. The observation of the SwinV2-L is similar to that of the SwinV2-B. With a larger image size $384\times 384$, MIM pre-trained SwinV2-L reaches $78.4$ on COCO $test$-$dev$, and $77.1$ on the challenging CrowdPose dataset. Using a stronger detection result from BigDetection~\cite{cai2022bigdet}, we obtain $80.5$ AP on COCO $val$, $78.9$ AP on COCO $test$-$dev$, and $78.0$ AP on CrowdPose.
	
	For the depth estimation, using a simple deconvolution head, SwinV2-B with MIM pre-training with ImageNet-1K achieves $0.304$ RMSE on NYUv2 and $2.050$ RMSE on KITTI, outperforming the previous SOTA method BinsFormer-L~\cite{li2022binsdepth}.
	The MIM pre-training does improve the performance of SwinV2-B by $0.03$ RMSE compared with the supervised pre-training with ImageNet-22K. Note that with supervised pre-training, a larger model SwinV2-L shows no gain for the NYUv2 dataset, while with MIM pre-training, SwinV2-L leads to about $0.02$ RMSE gain over SwinV2-B.
	
	For the video object tracking, MIM models also show a stronger transfer ability over supervised pre-trained models. On the long-term dataset LaSOT, SwinTrack~\cite{lin2021swintrack} with MIM pre-trained SwinV2-B backbone achieves comparable result with the SOTA MixFormer-L~\cite{cui2022mixfmr} with a larger image size $320\times 320$. We obtain the best SUC of $70.7$ on the LaSOT with SwinV2-L backbone with the input image size $224\times 224$ and template size $112\times 112$.

\subsection{Combined Task of Object Detection}

We select object detection on COCO as the combined task which simultaneously performs both semantic understanding and geometric learning. 
For object detection, a Mask-RCNN\cite{Mask-rcnn} framework is adopted and trained with a 3$\times$  schedule (36 epochs). We utilize an AdamW~\cite{kingma2014adam} optimizer with a learning rate of 6e-5/8e-5 for supervised/MIM model, a weight decay of 0.05, and a batch size of 32. We employ a large jittering augmentation (1024 $\times$ 1024 resolution, scale range [0.1, 2.0]).

\begin{figure}
    \centering
    \includegraphics[width=1.\linewidth]{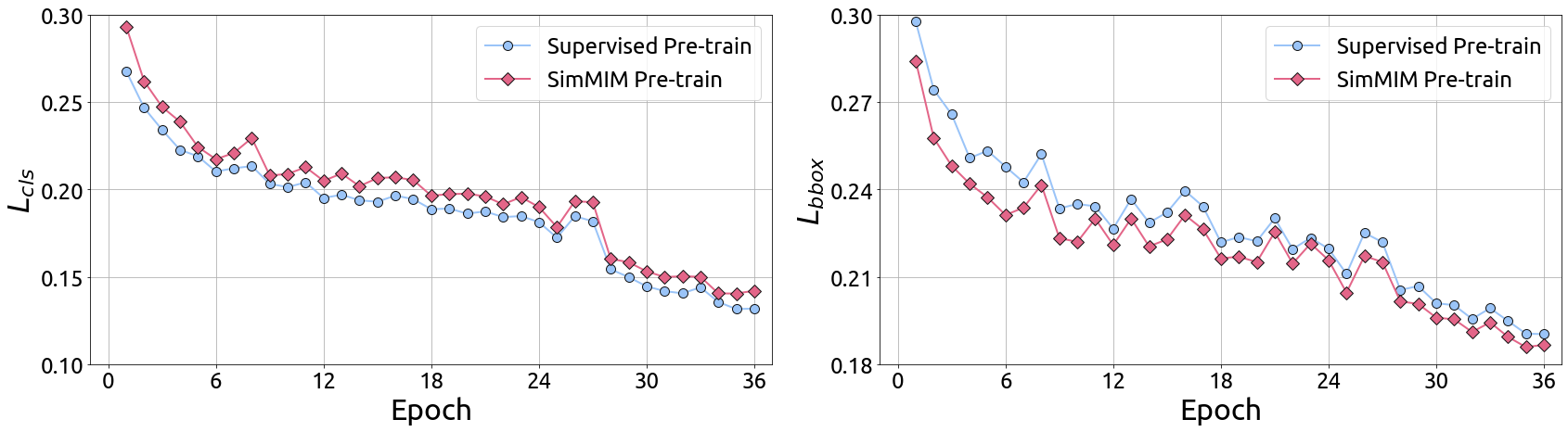}
    \vspace{-20pt}
    \caption{Loss curves of $L_{cls}$ and $L_{bbox}$ w.r.t the epoch number using supervised and MIM models with SwinV2-B as the backbone architecture.}
    \label{fig:det_loss}
\end{figure}

On COCO, we could clearly observe that MIM model outperforms its supervised counterpart (52.9/46.7 v.s. 51.9/45.7  of box/mask AP) with SwinV2-B as the backbone. We also plot the loss curves of object classification $L_{cls}$ and localization $L_{bbox}$, as shown in Figure~\ref{fig:det_loss}. We find that MIM model helps localization task converge faster and better, and the supervised model benefits more for object classification. This also matches our previous observations, that MIM model can perform better on geometric and motion tasks, and on par or slightly worse on the tasks that its categories are sufficiently covered by ImageNet like COCO. 

\section{Related Work}

\noindent\textbf{Visual Pre-training.} Throughout the deep learning era, supervised classification on ImageNet~\cite{deng2009imagenet} has been the dominant pretraining task. It is found to deliver strong finetuning performance on numerous semantic understanding tasks~\cite{donahue2014decaf,kornblith2019better,alex2019big,dosovitskiy2020vit,sermanet2013overfeat,girshick2014rich,liu2021swin,long2015fully,simonyan2014two,carreira2017quo}. Over the past several years, self-supervised pretraining has attracted more and more attention, and achieved finetuning performance on par with the supervised counterparts on several representative downstream tasks~\cite{he2019moco,chen2020simclr}, including two representative ones, contrastive learning~\cite{dosovitskiy2014exemplarcnn,he2019moco,chen2020simclr,cao2020pic,grill2020byol} and masked image modeling~\cite{chen2020imagegpt,bao2021beit,he2021masked,xie2021simmim}. 
In our work, we focus on understanding the different behaviors of supervised and emergent MIM pre-training.

\noindent\textbf{Understanding Pre-training.} There are some outstanding works~\cite{yosinski2014transferable,kornblith2019better,zhai2019large,nguyen2020wide,neyshabur2020being,raghu2021vision,park2022vision} trying to understand the pre-training procedure and inspire a lot of following works in a wide range. \cite{yosinski2014transferable} reveals how features of different layers are transferable in deep neural networks. \cite{kornblith2019better} performs a sufficient experimental study on different backbones and tries to answer whether better ImageNet models transfer better. Some works~\cite{raghu2021vision,zhou2021convnets,raghu2021vision} try to understand the behaviors of ViT, with CKA~\cite{cka}, loss landscape~\cite{li2018visualizing} and Fourier analysis.
In NLP, after BERT~\cite{devlin2018bert} pre-training came out, there is also a lot of works~\cite{kovaleva2019revealing,hao2019visualizing,clark2019does,hao2020self} trying to understand it. Most of them focus on the only interpretable component of Transformer, self-attention block, to give some detailed understanding.

\section{Conclusion}
In this work, we present a sufficient and sound analysis on masked image modeling, to reveal how and where MIM models work well. From visualizations, our most interesting finding is that the MIM pre-training brings locality to the trained model with sufficient diversity on the attention heads. This reveals why MIM is very helpful to the Vision Transformers (ViT, Swin, etc), because the Vision Transformer has a much larger receptive field, and to optimize it to a solution with strong generalization ability is difficult.
In experiments, our most interesting finding is that MIM pre-training can perform very well on the geometric and motion tasks with weak semantics. This finding helps the model to achieve state-of-the-art performance on those benchmarks without bells and whistles. 

It seems time to embrace masked image modeling as a general-purpose pre-trained model. We hope our paper can drive this belief deeper in the community and inspire new and solid research in this direction.
The best destination for an understanding paper would be to appear in the motivation of future technologies.

\bibliographystyle{apalike}
\bibliography{ref}

\begin{thebibliography}{}

\bibitem[Bahdanau et~al., 2014]{bahdanau2014neural}
Bahdanau, D., Cho, K., and Bengio, Y. (2014).
\newblock Neural machine translation by jointly learning to align and
  translate.
\newblock {\em arXiv preprint arXiv:1409.0473}.

\bibitem[Bao et~al., 2021]{bao2021beit}
Bao, H., Dong, L., and Wei, F. (2021).
\newblock Beit: Bert pre-training of image transformers.
\newblock {\em arXiv preprint arXiv:2106.08254}.

\bibitem[Cai et~al., 2022]{cai2022bigdet}
Cai, L., Zhang, Z., Zhu, Y., Zhang, L., Mu, L., and Xue, X. (2022).
\newblock Bigdetection: A large-scale benchmark for improved object detector
  pre-training.
\newblock {\em arXiv preprint arXiv:2203.13249}.

\bibitem[Cai and Vasconcelos, 2019]{cai2019cascadercnn}
Cai, Z. and Vasconcelos, N. (2019).
\newblock Cascade r-cnn: High quality object detection and instance
  segmentation.
\newblock {\em IEEE Transactions on Pattern Analysis and Machine Intelligence}.

\bibitem[Cao et~al., 2020]{cao2020pic}
Cao, Y., Xie, Z., Liu, B., Lin, Y., Zhang, Z., and Hu, H. (2020).
\newblock Parametric instance classification for unsupervised visual feature
  learning.
\newblock {\em Advances in Neural Information Processing Systems}, 33.

\bibitem[Carreira and Zisserman, 2017]{carreira2017quo}
Carreira, J. and Zisserman, A. (2017).
\newblock Quo vadis, action recognition? a new model and the kinetics dataset.
\newblock In {\em proceedings of the IEEE Conference on Computer Vision and
  Pattern Recognition}, pages 6299--6308.

\bibitem[Chen et~al., 2020a]{chen2020imagegpt}
Chen, M., Radford, A., Child, R., Wu, J., and Jun, H. (2020a).
\newblock Generative pretraining from pixels.
\newblock {\em Advances in Neural Information Processing Systems}.

\bibitem[Chen et~al., 2020b]{chen2020simclr}
Chen, T., Kornblith, S., Norouzi, M., and Hinton, G. (2020b).
\newblock A simple framework for contrastive learning of visual
  representations.
\newblock {\em ICML}.

\bibitem[Cheng et~al., 2021]{mask2former}
Cheng, B., Misra, I., Schwing, A.~G., Kirillov, A., and Girdhar, R. (2021).
\newblock Masked-attention mask transformer for universal image segmentation.
\newblock {\em arXiv preprint arXiv:2112.01527}.

\bibitem[Clark et~al., 2019]{clark2019does}
Clark, K., Khandelwal, U., Levy, O., and Manning, C.~D. (2019).
\newblock What does bert look at? an analysis of bert's attention.
\newblock {\em arXiv preprint arXiv:1906.04341}.

\bibitem[Cubuk et~al., 2018]{cubuk2018autoaugment}
Cubuk, E.~D., Zoph, B., Mane, D., Vasudevan, V., and Le, Q.~V. (2018).
\newblock Autoaugment: Learning augmentation policies from data.
\newblock {\em arXiv preprint arXiv:1805.09501}.

\bibitem[Cui et~al., 2022]{cui2022mixfmr}
Cui, Y., Jiang, C., Wang, L., and Wu, G. (2022).
\newblock Mixformer: End-to-end tracking with iterative mixed attention.
\newblock In {\em Proceedings of the IEEE/CVF conference on computer vision and
  pattern recognition}.

\bibitem[Deng et~al., 2009]{deng2009imagenet}
Deng, J., Dong, W., Socher, R., Li, L.-J., Li, K., and Fei-Fei, L. (2009).
\newblock Imagenet: A large-scale hierarchical image database.
\newblock In {\em CVPR}, pages 248--255. Ieee.

\bibitem[Devlin et~al., 2018]{devlin2018bert}
Devlin, J., Chang, M.-W., Lee, K., and Toutanova, K. (2018).
\newblock Bert: Pre-training of deep bidirectional transformers for language
  understanding.
\newblock {\em arXiv preprint arXiv:1810.04805}.

\bibitem[Ding et~al., 2022]{replknet}
Ding, X., Zhang, X., Zhou, Y., Han, J., Ding, G., and Sun, J. (2022).
\newblock Scaling up your kernels to 31x31: Revisiting large kernel design in
  cnns.
\newblock {\em arXiv preprint arXiv:2203.06717}.

\bibitem[Donahue et~al., 2014]{donahue2014decaf}
Donahue, J., Jia, Y., Vinyals, O., Hoffman, J., Zhang, N., Tzeng, E., and
  Darrell, T. (2014).
\newblock Decaf: A deep convolutional activation feature for generic visual
  recognition.
\newblock In {\em International conference on machine learning}, pages
  647--655. PMLR.

\bibitem[Dosovitskiy et~al., 2021]{dosovitskiy2020vit}
Dosovitskiy, A., Beyer, L., Kolesnikov, A., Weissenborn, D., Zhai, X.,
  Unterthiner, T., Dehghani, M., Minderer, M., Heigold, G., Gelly, S.,
  Uszkoreit, J., and Houlsby, N. (2021).
\newblock An image is worth 16x16 words: Transformers for image recognition at
  scale.
\newblock In {\em International Conference on Learning Representations}.

\bibitem[Dosovitskiy et~al., 2014]{dosovitskiy2014exemplarcnn}
Dosovitskiy, A., Springenberg, J.~T., Riedmiller, M., and Brox, T. (2014).
\newblock Discriminative unsupervised feature learning with convolutional
  neural networks.
\newblock In {\em Advances in neural information processing systems}, pages
  766--774.

\bibitem[Eigen et~al., 2014]{eigen2014depth}
Eigen, D., Puhrsch, C., and Fergus, R. (2014).
\newblock Depth map prediction from a single image using a multi-scale deep
  network.
\newblock In {\em Advances in neural information processing systems}, pages
  2366--2374.

\bibitem[Fan et~al., 2019]{fan2019lasot}
Fan, H., Lin, L., Yang, F., Chu, P., Deng, G., Yu, S., Bai, H., Xu, Y., Liao,
  C., and Ling, H. (2019).
\newblock Lasot: A high-quality benchmark for large-scale single object
  tracking.
\newblock In {\em Proceedings of the IEEE/CVF conference on computer vision and
  pattern recognition}, pages 5374--5383.

\bibitem[Fukushima, 1975]{fukushima1975cognitron}
Fukushima, K. (1975).
\newblock Cognitron: A self-organizing multilayered neural network.
\newblock {\em Biological cybernetics}, 20(3):121--136.

\bibitem[Geiger et~al., 2013]{andreas2013kitti}
Geiger, A., Lenz, P., Stiller, C., and Urtasun, R. (2013).
\newblock Vision meets robotics: The kitti dataset.
\newblock {\em International Journal of Robotics Research}, 32(11):1231--1237.

\bibitem[Geng et~al., 2021]{geng2021dekr}
Geng, Z., Sun, K., Xiao, B., Zhang, Z., and Wang, J. (2021).
\newblock Bottom-up human pose estimation via disentangled keypoint regression.
\newblock In {\em Proceedings of the IEEE/CVF conference on computer vision and
  pattern recognition}, pages 14676--14686.

\bibitem[Ghiasi et~al., 2021]{simple_copy_paste}
Ghiasi, G., Cui, Y., Srinivas, A., Qian, R., Lin, T.-Y., Cubuk, E.~D., Le,
  Q.~V., and Zoph, B. (2021).
\newblock Simple copy-paste is a strong data augmentation method for instance
  segmentation.
\newblock In {\em Proceedings of the IEEE/CVF Conference on Computer Vision and
  Pattern Recognition (CVPR)}, pages 2918--2928.

\bibitem[Girshick et~al., 2014]{girshick2014rich}
Girshick, R., Donahue, J., Darrell, T., and Malik, J. (2014).
\newblock Rich feature hierarchies for accurate object detection and semantic
  segmentation.
\newblock In {\em Proceedings of the IEEE conference on computer vision and
  pattern recognition}, pages 580--587.

\bibitem[Gretton et~al., 2005]{hsic}
Gretton, A., Bousquet, O., Smola, A., and Sch{\"o}lkopf, B. (2005).
\newblock Measuring statistical dependence with hilbert-schmidt norms.
\newblock In {\em International conference on algorithmic learning theory},
  pages 63--77. Springer.

\bibitem[Grill et~al., 2020]{grill2020byol}
Grill, J.-B., Strub, F., Altch{\'e}, F., Tallec, C., Richemond, P.,
  Buchatskaya, E., Doersch, C., Avila~Pires, B., Guo, Z., Gheshlaghi~Azar, M.,
  et~al. (2020).
\newblock Bootstrap your own latent-a new approach to self-supervised learning.
\newblock {\em Advances in Neural Information Processing Systems}, 33.

\bibitem[Hao et~al., 2019]{hao2019visualizing}
Hao, Y., Dong, L., Wei, F., and Xu, K. (2019).
\newblock Visualizing and understanding the effectiveness of bert.
\newblock {\em arXiv preprint arXiv:1908.05620}.

\bibitem[Hao et~al., 2020]{hao2020self}
Hao, Y., Dong, L., Wei, F., and Xu, K. (2020).
\newblock Self-attention attribution: Interpreting information interactions
  inside transformer.
\newblock {\em arXiv preprint arXiv:2004.11207}, 2.

\bibitem[He et~al., 2021]{he2021masked}
He, K., Chen, X., Xie, S., Li, Y., Doll{\'a}r, P., and Girshick, R. (2021).
\newblock Masked autoencoders are scalable vision learners.
\newblock {\em arXiv preprint arXiv:2111.06377}.

\bibitem[He et~al., 2020]{he2019moco}
He, K., Fan, H., Wu, Y., Xie, S., and Girshick, R. (2020).
\newblock Momentum contrast for unsupervised visual representation learning.
\newblock {\em CVPR}.

\bibitem[He et~al., 2017]{Mask-rcnn}
He, K., Gkioxari, G., Doll{\'a}r, P., and Girshick, R. (2017).
\newblock Mask r-cnn.
\newblock In {\em ICCV}, pages 2961--2969.

\bibitem[He et~al., 2016]{he2016resnet}
He, K., Zhang, X., Ren, S., and Sun, J. (2016).
\newblock Deep residual learning for image recognition.
\newblock In {\em CVPR}, pages 770--778.

\bibitem[Huang et~al., 2016]{huang2016deep}
Huang, G., Sun, Y., Liu, Z., Sedra, D., and Weinberger, K.~Q. (2016).
\newblock Deep networks with stochastic depth.
\newblock In {\em European conference on computer vision}, pages 646--661.
  Springer.

\bibitem[Huang et~al., 2020]{huang2020udp}
Huang, J., Zhu, Z., Guo, F., and Huang, G. (2020).
\newblock The devil is in the details: Delving into unbiased data processing
  for human pose estimation.
\newblock In {\em Proceedings of the IEEE/CVF conference on computer vision and
  pattern recognition}, pages 5699--5708.

\bibitem[Huang et~al., 2021]{huang2021got10k}
Huang, L., Zhao, X., and Huang, K. (2021).
\newblock Got-10k: A large high-diversity benchmark for generic object tracking
  in the wild.
\newblock {\em IEEE transactions on pattern analysis and machine intelligence},
  43(5):1562--1577.

\bibitem[Hubel and Wiesel, 1962]{hubel1962receptive}
Hubel, D.~H. and Wiesel, T.~N. (1962).
\newblock Receptive fields, binocular interaction and functional architecture
  in the cat's visual cortex.
\newblock {\em The Journal of physiology}, 160(1):106--154.

\bibitem[Kim et~al., 2022]{kim2022glpdepth}
Kim, D., Ga, W., Ahn, P., Joo, D., Chun, S., and Kim, J. (2022).
\newblock Global-local path networks for monocular depth estimation with
  vertical cutdepth.
\newblock {\em arXiv preprint arXiv:2201.07436}.

\bibitem[Kingma and Ba, 2014]{kingma2014adam}
Kingma, D.~P. and Ba, J. (2014).
\newblock Adam: A method for stochastic optimization.
\newblock {\em arXiv preprint arXiv:1412.6980}.

\bibitem[Kolesnikov et~al., 2019]{alex2019big}
Kolesnikov, A., Beyer, L., Zhai, X., Puigcerver, J., Yung, J., Gelly, S., and
  Houlsby, N. (2019).
\newblock Big transfer (bit): General visual representation learning.

\bibitem[Kornblith et~al., 2019a]{cka}
Kornblith, S., Norouzi, M., Lee, H., and Hinton, G. (2019a).
\newblock Similarity of neural network representations revisited.
\newblock In {\em International Conference on Machine Learning}, pages
  3519--3529. PMLR.

\bibitem[Kornblith et~al., 2019b]{kornblith2019better}
Kornblith, S., Shlens, J., and Le, Q.~V. (2019b).
\newblock Do better imagenet models transfer better?
\newblock In {\em Proceedings of the IEEE/CVF conference on computer vision and
  pattern recognition}, pages 2661--2671.

\bibitem[Kovaleva et~al., 2019]{kovaleva2019revealing}
Kovaleva, O., Romanov, A., Rogers, A., and Rumshisky, A. (2019).
\newblock Revealing the dark secrets of bert.
\newblock {\em arXiv preprint arXiv:1908.08593}.

\bibitem[Krizhevsky et~al., 2012]{alexnet}
Krizhevsky, A., Sutskever, I., and Hinton, G.~E. (2012).
\newblock Imagenet classification with deep convolutional neural networks.
\newblock In {\em Advances in Neural Information Processing Systems}, pages
  1097--1105.

\bibitem[Kullback and Leibler, 1951]{kldivergence}
Kullback, S. and Leibler, R.~A. (1951).
\newblock On information and sufficiency.
\newblock {\em The annals of mathematical statistics}, 22(1):79--86.

\bibitem[LeCun et~al., 1999]{lecun1999object}
LeCun, Y., Haffner, P., Bottou, L., and Bengio, Y. (1999).
\newblock Object recognition with gradient-based learning.
\newblock In {\em Shape, contour and grouping in computer vision}, pages
  319--345. Springer.

\bibitem[Li et~al., 2018]{li2018visualizing}
Li, H., Xu, Z., Taylor, G., Studer, C., and Goldstein, T. (2018).
\newblock Visualizing the loss landscape of neural nets.
\newblock {\em Advances in neural information processing systems}, 31.

\bibitem[Li et~al., 2019]{li2019crowd}
Li, J., Wang, C., Zhu, H., Mao, Y., Fang, H., and Lu, C. (2019).
\newblock Crowdpose: Efficient crowded scenes pose estimation and a new
  benchmark.
\newblock In {\em Proceedings of the IEEE/CVF conference on computer vision and
  pattern recognition}, pages 10863--10872.

\bibitem[Li et~al., 2021]{li2021localvit}
Li, Y., Zhang, K., Cao, J., Timofte, R., and Van~Gool, L. (2021).
\newblock Localvit: Bringing locality to vision transformers.
\newblock {\em arXiv preprint arXiv:2104.05707}.

\bibitem[Li et~al., 2022]{li2022binsdepth}
Li, Z., Wang, X., Liu, X., and Jiang, J. (2022).
\newblock Binsformer: Revisiting adaptive bins for monocular depth estimation.
\newblock {\em arXiv preprint arXiv:2204.00987}.

\bibitem[Lin et~al., 2021]{lin2021swintrack}
Lin, L., Fan, H., Xu, Y., and Ling, H. (2021).
\newblock Swintrack: A simple and strong baseline for transformer tracking.
\newblock {\em arXiv preprint arXiv:2112.00995}.

\bibitem[Lin et~al., 2014]{lin2014microsoft}
Lin, T.-Y., Maire, M., Belongie, S., Hays, J., Perona, P., Ramanan, D.,
  Doll{\'a}r, P., and Zitnick, C.~L. (2014).
\newblock Microsoft coco: Common objects in context.
\newblock In {\em ECCV}, pages 740--755. Springer.

\bibitem[Liu et~al., 2019]{liu2019roberta}
Liu, Y., Ott, M., Goyal, N., Du, J., Joshi, M., Chen, D., Levy, O., Lewis, M.,
  Zettlemoyer, L., and Stoyanov, V. (2019).
\newblock Roberta: A robustly optimized bert pretraining approach.
\newblock {\em arXiv preprint arXiv:1907.11692}.

\bibitem[Liu et~al., 2021a]{swinv2}
Liu, Z., Hu, H., Lin, Y., Yao, Z., Xie, Z., Wei, Y., Ning, J., Cao, Y., Zhang,
  Z., Dong, L., et~al. (2021a).
\newblock Swin transformer v2: Scaling up capacity and resolution.
\newblock {\em arXiv preprint arXiv:2111.09883}.

\bibitem[Liu et~al., 2021b]{liu2021swin}
Liu, Z., Lin, Y., Cao, Y., Hu, H., Wei, Y., Zhang, Z., Lin, S., and Guo, B.
  (2021b).
\newblock Swin transformer: Hierarchical vision transformer using shifted
  windows.
\newblock {\em arXiv preprint arXiv:2103.14030}.

\bibitem[Liu et~al., 2021c]{liu2021video}
Liu, Z., Ning, J., Cao, Y., Wei, Y., Zhang, Z., Lin, S., and Hu, H. (2021c).
\newblock Video swin transformer.

\bibitem[Long et~al., 2015]{long2015fully}
Long, J., Shelhamer, E., and Darrell, T. (2015).
\newblock Fully convolutional networks for semantic segmentation.
\newblock In {\em Proceedings of the IEEE conference on computer vision and
  pattern recognition}, pages 3431--3440.

\bibitem[Loshchilov and Hutter, 2019]{Loshchilov2019adamw}
Loshchilov, I. and Hutter, F. (2019).
\newblock Decoupled weight decay regularization.
\newblock In {\em International Conference on Learning Representations}.

\bibitem[Müller et~al., 2018]{muller2018tracknet}
Müller, M., Bibi, A., Giancola, S., Al-Subaihi, S., and Ghanem, B. (2018).
\newblock Trackingnet: A large-scale dataset and benchmark for object tracking
  in the wild.
\newblock In {\em Proceedings of the European Conference on Computer Vision},
  pages 310--327. Springer.

\bibitem[Neyshabur et~al., 2020]{neyshabur2020being}
Neyshabur, B., Sedghi, H., and Zhang, C. (2020).
\newblock What is being transferred in transfer learning?
\newblock {\em Advances in neural information processing systems}, 33:512--523.

\bibitem[Nguyen et~al., 2020]{nguyen2020wide}
Nguyen, T., Raghu, M., and Kornblith, S. (2020).
\newblock Do wide and deep networks learn the same things? uncovering how
  neural network representations vary with width and depth.
\newblock {\em arXiv preprint arXiv:2010.15327}.

\bibitem[Park and Kim, 2022]{park2022vision}
Park, N. and Kim, S. (2022).
\newblock How do vision transformers work?
\newblock {\em arXiv preprint arXiv:2202.06709}.

\bibitem[Raghu et~al., 2021]{raghu2021vision}
Raghu, M., Unterthiner, T., Kornblith, S., Zhang, C., and Dosovitskiy, A.
  (2021).
\newblock Do vision transformers see like convolutional neural networks?
\newblock {\em Advances in Neural Information Processing Systems}, 34.

\bibitem[Ranftl et~al., 2021]{ranftl2021dpt}
Ranftl, R., Bochkovskiy, A., and Koltun, V. (2021).
\newblock Vision transformers for dense prediction.
\newblock In {\em Proceedings of the IEEE/CVF International Conference on
  Computer Vision}, pages 12159--12168.

\bibitem[Ren et~al., 2015]{ren2015faster}
Ren, S., He, K., Girshick, R., and Sun, J. (2015).
\newblock Faster r-cnn: Towards real-time object detection with region proposal
  networks.
\newblock In {\em Advances in neural information processing systems}, pages
  91--99.

\bibitem[Sariyildiz et~al., 2021]{sariyildiz2021cog}
Sariyildiz, M.~B., Kalantidis, Y., Larlus, D., and Alahari, K. (2021).
\newblock Concept generalization in visual representation learning.
\newblock In {\em Proceedings of the IEEE/CVF International Conference on
  Computer Vision}, pages 9629--9639.

\bibitem[Sermanet et~al., 2013]{sermanet2013overfeat}
Sermanet, P., Eigen, D., Zhang, X., Mathieu, M., Fergus, R., and LeCun, Y.
  (2013).
\newblock Overfeat: Integrated recognition, localization and detection using
  convolutional networks.
\newblock {\em arXiv preprint arXiv:1312.6229}.

\bibitem[Silberman et~al., 2012]{nathan2012nyuv2}
Silberman, N., Hoiem, D., Kohli, P., and Fergus, R. (2012).
\newblock Indoor segmentation and support inference from rgbd images.
\newblock In {\em Proceedings of the European Conference on Computer Vision},
  pages 746--760. Springer.

\bibitem[Simonyan and Zisserman, 2014]{simonyan2014two}
Simonyan, K. and Zisserman, A. (2014).
\newblock Two-stream convolutional networks for action recognition in videos.
\newblock In {\em Advances in neural information processing systems}, pages
  568--576.

\bibitem[Sun et~al., 2019]{sun2019hrnet}
Sun, K., Xiao, B., Liu, D., and Wang, J. (2019).
\newblock Deep high-resolution representation learning for human pose
  estimation.
\newblock In {\em Proceedings of the IEEE/CVF conference on computer vision and
  pattern recognition}, pages 5693--5703.

\bibitem[Szegedy et~al., 2016]{szegedy2016rethinking}
Szegedy, C., Vanhoucke, V., Ioffe, S., Shlens, J., and Wojna, Z. (2016).
\newblock Rethinking the inception architecture for computer vision.
\newblock In {\em Proceedings of the IEEE conference on computer vision and
  pattern recognition}, pages 2818--2826.

\bibitem[Tran et~al., 2015]{tran2015learning}
Tran, D., Bourdev, L., Fergus, R., Torresani, L., and Paluri, M. (2015).
\newblock Learning spatiotemporal features with 3d convolutional networks.
\newblock In {\em Proceedings of the IEEE international conference on computer
  vision}, pages 4489--4497.

\bibitem[Van~Horn et~al., 2018]{van2018inaturalist}
Van~Horn, G., Mac~Aodha, O., Song, Y., Cui, Y., Sun, C., Shepard, A., Adam, H.,
  Perona, P., and Belongie, S. (2018).
\newblock The inaturalist species classification and detection dataset.
\newblock In {\em Proceedings of the IEEE conference on computer vision and
  pattern recognition}, pages 8769--8778.

\bibitem[Wang et~al., 2018]{wang2018non}
Wang, X., Girshick, R., Gupta, A., and He, K. (2018).
\newblock Non-local neural networks.
\newblock In {\em Proceedings of the IEEE conference on computer vision and
  pattern recognition}, pages 7794--7803.

\bibitem[Xiao et~al., 2018a]{xiao2018sbaseline}
Xiao, B., Wu, H., and Wei, Y. (2018a).
\newblock Simple baselines for human pose estimation and tracking.
\newblock In {\em Proceedings of the European Conference on Computer Vision},
  pages 472--487. Springer.

\bibitem[Xiao et~al., 2018b]{xiao2018upernet}
Xiao, T., Liu, Y., Zhou, B., Jiang, Y., and Sun, J. (2018b).
\newblock Unified perceptual parsing for scene understanding.
\newblock In {\em Proceedings of the European Conference on Computer Vision
  (ECCV)}, pages 418--434.

\bibitem[Xie et~al., 2021]{xie2021simmim}
Xie, Z., Zhang, Z., Cao, Y., Lin, Y., Bao, J., Yao, Z., Dai, Q., and Hu, H.
  (2021).
\newblock Simmim: A simple framework for masked image modeling.
\newblock {\em arXiv preprint arXiv:2111.09886}.

\bibitem[Yosinski et~al., 2014]{yosinski2014transferable}
Yosinski, J., Clune, J., Bengio, Y., and Lipson, H. (2014).
\newblock How transferable are features in deep neural networks?
\newblock {\em Advances in neural information processing systems}, 27.

\bibitem[Yuan et~al., 2021]{yuan2021hrformer}
Yuan, Y., Fu, R., Huang, L., Lin, W., Zhang, C., Chen, X., and Wang, J. (2021).
\newblock Hrformer: High-resolution vision transformer for dense predict.
\newblock In {\em Advances in neural information processing systems}, pages
  7281--7293.

\bibitem[Yun et~al., 2019]{yun2019cutmix}
Yun, S., Han, D., Oh, S.~J., Chun, S., Choe, J., and Yoo, Y. (2019).
\newblock Cutmix: Regularization strategy to train strong classifiers with
  localizable features.
\newblock In {\em Proceedings of the IEEE/CVF International Conference on
  Computer Vision}, pages 6023--6032.

\bibitem[Zhai et~al., 2019]{zhai2019large}
Zhai, X., Puigcerver, J., Kolesnikov, A., Ruyssen, P., Riquelme, C., Lucic, M.,
  Djolonga, J., Pinto, A.~S., Neumann, M., Dosovitskiy, A., et~al. (2019).
\newblock A large-scale study of representation learning with the visual task
  adaptation benchmark.
\newblock {\em arXiv preprint arXiv:1910.04867}.

\bibitem[Zhang et~al., 2017]{zhang2017mixup}
Zhang, H., Cisse, M., Dauphin, Y.~N., and Lopez-Paz, D. (2017).
\newblock mixup: Beyond empirical risk minimization.
\newblock {\em arXiv preprint arXiv:1710.09412}.

\bibitem[Zhong et~al., 2020]{zhong2020random}
Zhong, Z., Zheng, L., Kang, G., Li, S., and Yang, Y. (2020).
\newblock Random erasing data augmentation.
\newblock In {\em Proceedings of the AAAI Conference on Artificial
  Intelligence}, volume~34, pages 13001--13008.

\bibitem[Zhou et~al., 2021a]{zhou2021convnets}
Zhou, H.-Y., Lu, C., Yang, S., and Yu, Y. (2021a).
\newblock Convnets vs. transformers: Whose visual representations are more
  transferable?
\newblock In {\em Proceedings of the IEEE/CVF International Conference on
  Computer Vision}, pages 2230--2238.

\bibitem[Zhou et~al., 2021b]{zhou2021ibot}
Zhou, J., Wei, C., Wang, H., Shen, W., Xie, C., Yuille, A., and Kong, T.
  (2021b).
\newblock ibot: Image bert pre-training with online tokenizer.
\newblock {\em arXiv preprint arXiv:2111.07832}.

\end{thebibliography}
\newpage

\appendix

\section{Visualizations on Swin Transformer}\label{sec:vis_swin}

It is crucial to know whether our observations in visualizations are general across different backbone architectures. Thanks to the general applicability of SimMIM~\cite{xie2021simmim}, we further perform the visualizations on SwinV2-B~\cite{swinv2} (in Section~\ref{sec:vis_swin}) and RepLKNet~\cite{replknet} (in Section~\ref{sec:lknet}). Fortunately, we find that most of the observations could be transferred across architectures, ViT-B, SwinV2-B, and RepLKNet.

\subsection{Visualizations on Attention Maps}

\begin{figure}[h]
    \centering
    \includegraphics[width=1.\linewidth]{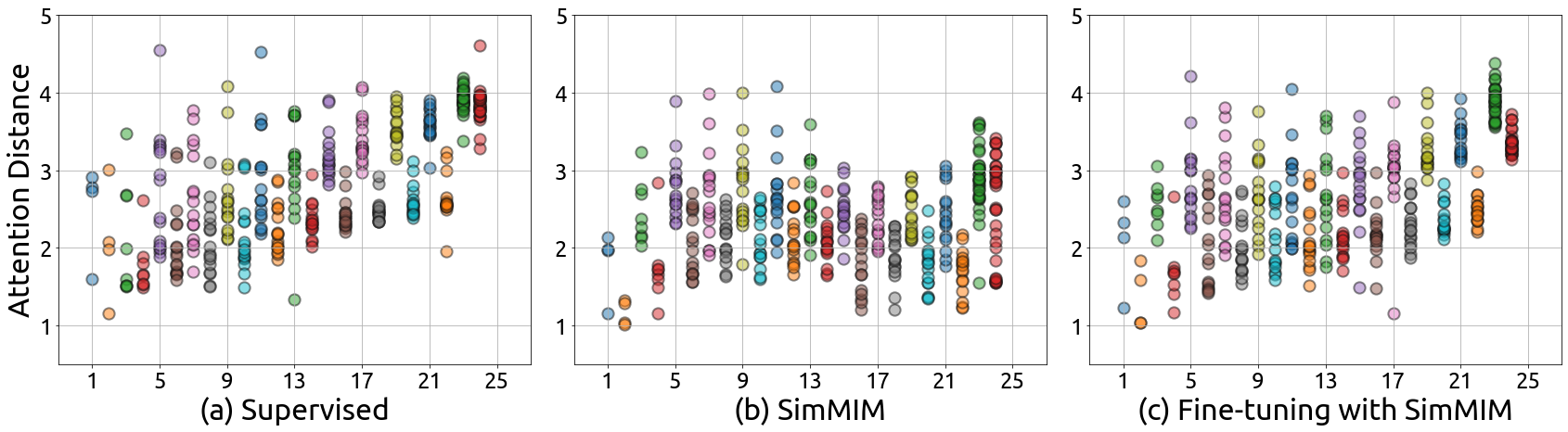}
  \vspace{-2.0em}
    \caption{The averaged attention distance in different attention heads (dots) w.r.t the layer number on (a) supervised model, (b) SimMIM model, and (c) supervised fine-tuned model with SimMIM pre-training with SwinV2-B as the backbone.}
    \label{fig:swin_attention_distance}
\end{figure}

\paragraph{Local Attention or Global Attention?} Results are shown in  Figure~\ref{fig:swin_attention_distance}.
First, we can have a similar observation as in ViT-B that the supervised model (a) tends to focus locally at lower layers but more globally at higher layers, and the SimMIM model (b) tends to aggregate both local and global pixels in all layers, and the average attention distance of SimMIM model is similar to the lower layers of the supervised counterpart.
The supervised fine-tuned model (c) with SimMIM pre-training behaves very similarly to the supervised model trained from scratch, but still maintains some good properties in SimMIM pre-training (a larger diversity on the last several layers). 
Also, we find that the averaged aggregated distances in two consecutive layers are one high and one low. This is due to the shifted windowing scheme in Swin Transformer, that is, the ranges that each pixel can aggregate in two consecutive layers are different.

\begin{figure}[h]
    \centering
    \includegraphics[width=1.\linewidth]{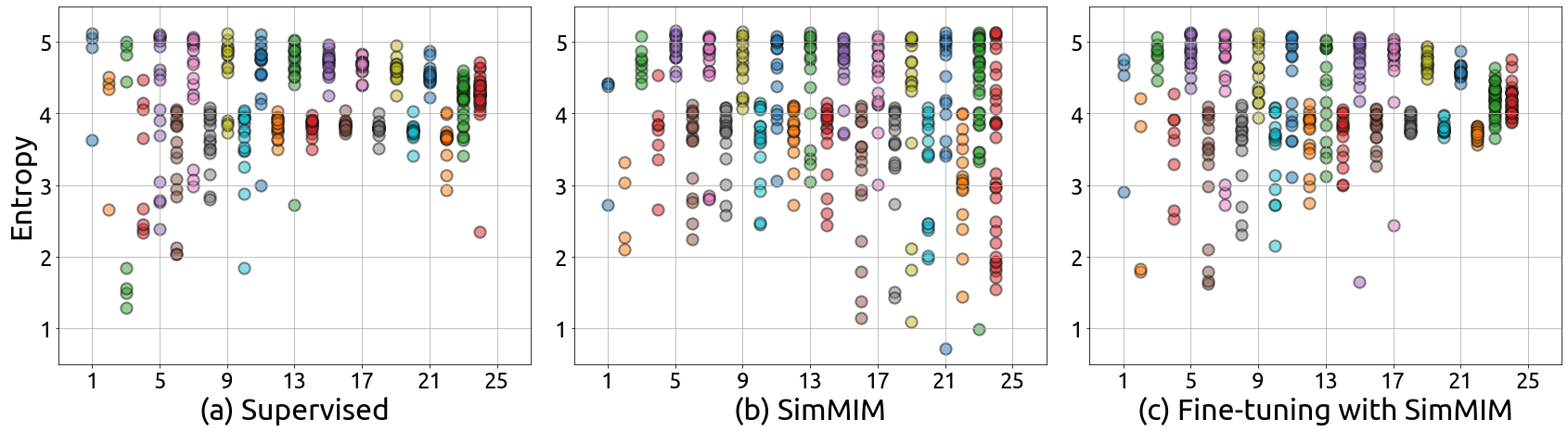}
  \vspace{-2.0em}
    \caption{The entropy of each head’s attention distribution in different attention heads (dots) w.r.t the layer number on (a) supervised model, (b) SimMIM model, and (c) supervised fine-tuned model with SimMIM pre-training with SwinV2-B as the backbone.}
    \label{fig:swin_entropy}
\end{figure}

\paragraph{Focused Attention or Broad Attention?} A similar observation could be found with Swin-B as the backbone as using ViT-B as the backbone in the main paper, as shown in Figure~\ref{fig:swin_entropy}. 

\begin{figure}[h]
    \centering
    \includegraphics[width=1.\linewidth]{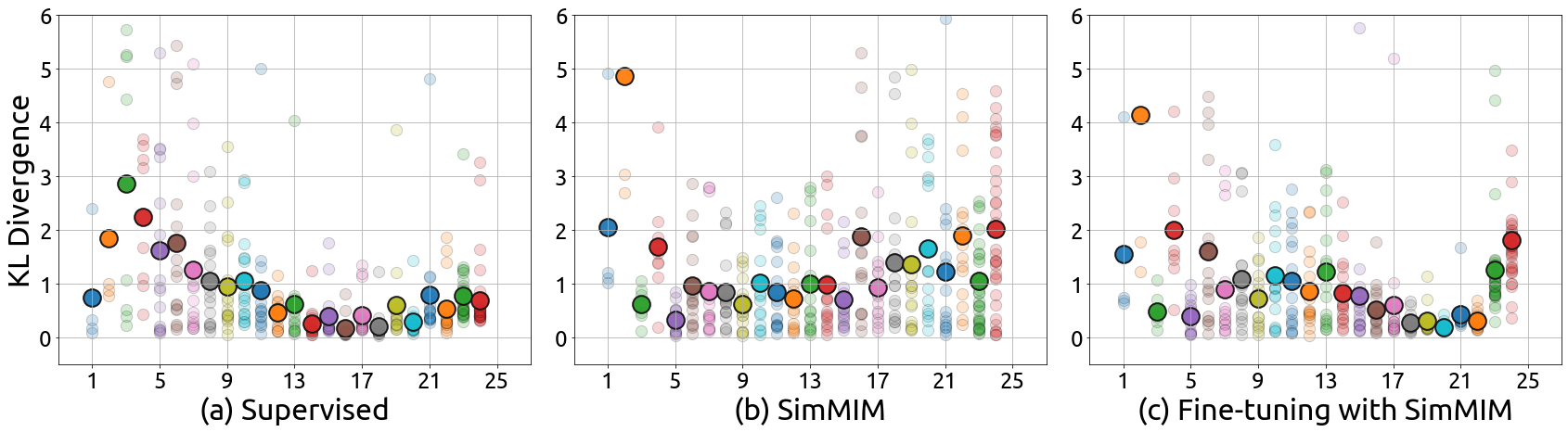}
  \vspace{-2.0em}
    \caption{The KL divergence between attention distributions of different heads (small dots) and the averaged KL divergence (large dots) in each layer w.r.t the layer number on (a) supervised model, (b) SimMIM model, and (c) supervised fine-tuned model with SimMIM pre-training with SwinV2-B as the backbone.}
    \label{fig:swin_kl_divergence}
\end{figure}

\paragraph{Diversity on Attention Heads} As shown in Figure~\ref{fig:swin_kl_divergence}, similar to ViT-B, in SimMIM models (b), different attention heads tend to aggregate different tokens on all layers. But for supervised models (a), the diversity on attention heads becomes smaller as the layer goes deeper. Interestingly, after supervised fine-tuning the SimMIM model on ImageNet-1K, the model (c) behaves much more similarly to the supervised model (a) trained from scratch, but maintains an advantage of the SimMIM model, that is, a larger diversity on attention heads of the last two layers.

\subsection{Investigating the Representation Structures via CKA Similarity}

\begin{figure}[h]
    \centering
    \includegraphics[width=1.\linewidth]{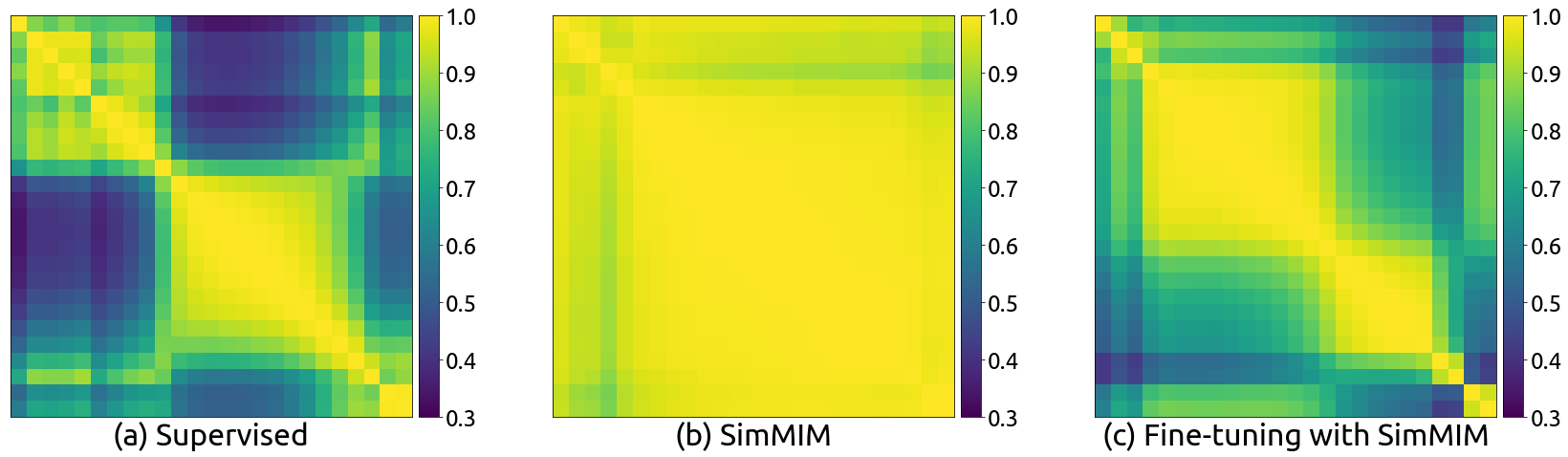}
  \vspace{-2.0em}
    \caption{The CKA heatmap between the feature maps of different layers of (a) supervised model, (b) SimMIM model, and (c) supervised fine-tuned model with SimMIM pre-training with SwinV2-B as the backbone.}
    \label{fig:swin_cka}
\end{figure}

It is challenging to analyze and compare the layer representations of deep networks, because their features are high-dimensional and with different dimensions. Centered kernel alignment (CKA)~\cite{cka} is defined to address this challenge, and enables quantitative comparisons of feature representations within and across networks. Given two inputs of ${ X} \in \mathbb{R}^{N\times D_1}$ and ${ Y} \in \mathbb{R}^{N\times D_2}$, where $N$ denotes number of examples and $D_1$ and $D_2$ denote the dimension. Then the Gram matrices are computed as $K=X X^T$ and $L=Y  Y^T$. CKA is then defined as
\begin{equation}\label{eqn:cka}
    \text{CKA}(K, L) = \frac{\text{HSIC}(K, L)}{\sqrt{\text{HSIC}(K, K) \text{HSIC}(L, L)}},
\end{equation}
where HSIC($\cdot$,$\cdot$) denotes the Hilbert-Schmidt independence criterion~\cite{hsic}. Note that, CKA is invariant to the orthogonal transformation and isotropic scaling, which enables valuable and effective comparison and analysis on hidden representations of deep networks. 

Results of CKA similarity between feature representations of different layers on (a) supervised model, (b) SimMIM model, and (c) supervised fine-tuned model with SimMIM pre-training with SwinV2-B as the backbone, are shown in Figure~\ref{fig:swin_cka}. We still have a similar observation as in ViT-B, that the representation structures of different layers in SimMIM models are almost the same, and supervised models trained from scratch learn different representation structures in different layers. With the help of the SimMIM pre-training, the representation structures of different layers in supervised model are not as different as that in the scratch supervised models.

\section{Investigations on Large-kernel ConvNets (RepLKNet~\cite{replknet})}\label{sec:lknet}

From the previous visualizations on Vision Transformers (ViT) and Swin Transformers, we find that the MIM pre-training brings the locality inductive bias and larger diversity on attention heads to the trained models comparing to the supervised counterpart, which may benefit the optimization of the trained models on downstream tasks.  This reminds us that large-kernel ConvNets~\cite{replknet} without special designs still face the optimization issue, and need the re-parametrization trick with small kernels to bring the locality back and help them optimize. Thus it is valuable to know whether the masked image modeling (MIM) as pre-training could help the large-kernel ConvNets to optimize without the re-parametrization trick.
Thanks to the general applicability of SimMIM~\cite{xie2021simmim}, we could also perform experiments and visualizations on large-kernel ConvNets~\cite{replknet} with the MIM pre-training.

\subsection{Experimental Results}

\renewcommand{\arraystretch}{1.12}
\begin{table}[h]
  \centering
  \addtolength{\tabcolsep}{-1.5pt}
  \begin{tabular}{m{3.02cm}<{\centering}m{3.5cm}<{\centering}|c|m{1.19cm}<{\centering}m{1.19cm}<{\centering}m{1.19cm}<{\centering}}
    \bottomrule
    \multirow{3}{*}{backbone}  & \multirow{3}{*}{pre-train}  &  \multirow{3}{*}{ImageNet-1K} &  \multicolumn{3}{c}{Pose Estimation} \\
    & & & COCO & COCO & Crowd-  \\
    & & &  $val$ & $test$ & Pose\\
    \hline
    RepLKNet-31B & 1K-SUP w/ Reparam. & $83.5$ & $74.6$ & $73.9$ & $70.2$\\
    RepLKNet-31B & 1K-MIM w/o Reparam. & $83.3$ & $76.5$ & $75.8$ & $72.4$\\
    \toprule
  \end{tabular}
  \caption{Detailed comparisons of pre-trained RepLKNet models on the classification and the pose estimation tasks. We report the top-1 accuracy ($\uparrow$) for the ImageNet-1K dataset and the AP ($\uparrow$) for the pose estimation tasks.
  }
  \label{tab-lknet}
  \vspace{-1.0em}
\end{table}

\paragraph{Setup}
For MIM pretraining, we utilize the RepLKNet-31B~\cite{replknet} without the specially designed re-parametrization trick. Before the stem of the RepLKNet, using a normal $1\times 1$ convolution, we map the 3-dimension space of the image into a high-dimensional space where we randomly mask out some patches. Following SimMIM~\cite{xie2021simmim}, the image size is $192\times 192$, we divide it into $6\times 6$ patches and randomly mask out $60\%$ patches. The decoder contains a linear projection layer and an upsample layer. We use $\ell_1$-loss to supervise the reconstruction of the masked pixels.

We use the ImageNet-1k for MIM pre-training and augment the data using the random resize cropping (scale range $[0.67, 1]$ and aspect ratio range $[3/4, 4/3]$), and random flipping. The optimizer is the AdamW\cite{Loshchilov2019adamw} optimizer with a weight decay of $5e$-$2$ and a base learning rate of $4e$-$4$. We use warm-up for 10 epochs, drop the learning rate to $4e$-5 at $260$th epoch, and train for $300$ epochs in total. The batch size is $2048$. We use the DropPath of $0.1$ for RepLKNet-31B and gradient clipping.

We report the top-$1$ accuracy of the supervised pre-trained model on ImageNet-1k in the original paper~\cite{replknet}. For fine-tuning of MIM pre-trained model on ImageNet-1k, we follow the setting of SimMIM~\cite{xie2021simmim} and use the AdamW optimizer with a weight decay of $5e$-$2$, a base learning rate of $5e$-$3$ with a layer decay of $0.8$. The learning rate is scheduled via cosine strategy and we use 20 epochs for warm-up and train for $100$ epochs in total. The batch size is $2048$. We adopt the DropPath of $0.1$ and gradient clipping. The data augmentations contain AutoAug~\cite{cubuk2018autoaugment}, Mixup~\cite{zhang2017mixup}, CutMix~\cite{yun2019cutmix}, color jitter, random erasing~\cite{zhong2020random}, and label smoothing~\cite{szegedy2016rethinking}.
The settings of the pose estimation are the same as the details in Section~\ref{sec:exp-detail}.

\paragraph{Results} As shown in Table~\ref{tab-lknet}, the MIM pre-training can help the large-kernel convnets to address the optimization issue to some extent and achieve on par performance on ImageNet-1K compared with the supervised model with the re-parametrization trick. Note that, on pose estimation, MIM models still surpass supervised counterparts with the re-parametrization trick by large margins, which indicates that the benefit of MIM pre-training on geometric and motion tasks is general across different backbone architectures.

\subsection{Visualizations} 

To further understand whether the behaviors of large-kernel ConvNets with MIM pre-training are similar to those of Vision/Swin Transformers, we visualize the convolutional kernels with similar tools used in visualizing the attention maps. As the basic component in RepLKNet is the depth-wise convolution with the kernel dimension of $C\times H\times W$, we normalize each channel of the depth-wise convolutional kernels (on the dimension of $H\times W$) to make them as a similar role of attention map, and regard different channels ($C$ channels) of the depth-wise convolutional kernels as the attention heads. Then we could directly apply the previous tools on attention maps for visualizations.

\begin{figure}[h]
    \centering
    \includegraphics[width=1.\linewidth]{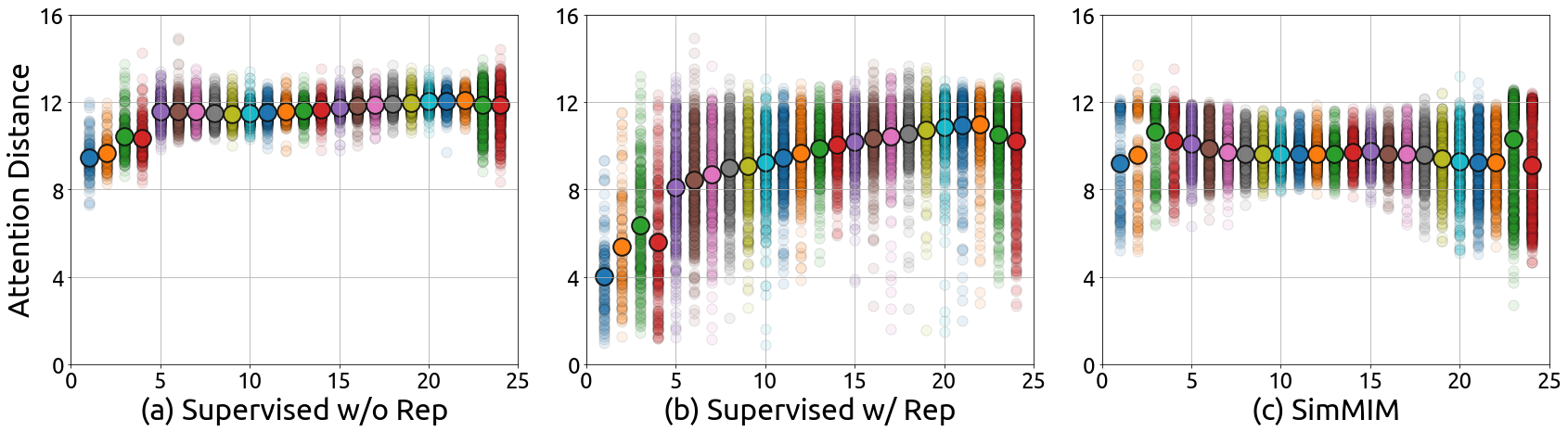}
    \caption{The aggregated distance in different channels (small dots) and the averaged aggregated distance (large dots) w.r.t the layer number on (a) supervised model without the re-parametrization trick, (b) supervised model with the re-parametrization trick, and (c) SimMIM model, with RepLKNet-31B as the backbone architecture.}
    \label{fig:replknet_attention_distance}
\end{figure}

\paragraph{Local Kernels or Global Kernels?}

As shown in  Figure~\ref{fig:replknet_attention_distance}, with the re-parametrization trick, the RepLKNet-31B model (b) with supervised training focuses much more locally in all layers. Similar to previous supervised trained models, RepLKNet-31B models with supervised training still tend to focus locally at lower layers but more globally at higher layers. But for the model trained by SimMIM (c), each layer has diverse kernels that tend to aggregate both local and global pixels, and the average aggregated distance is much smaller than the supervised trained model without the re-parametrization trick (a), indicating that MIM still brings locality inductive bias to the large-kernel ConvNets with a similar role of the re-parametrization trick but less strength.

\begin{figure}[h]
    \centering
    \includegraphics[width=1.\linewidth]{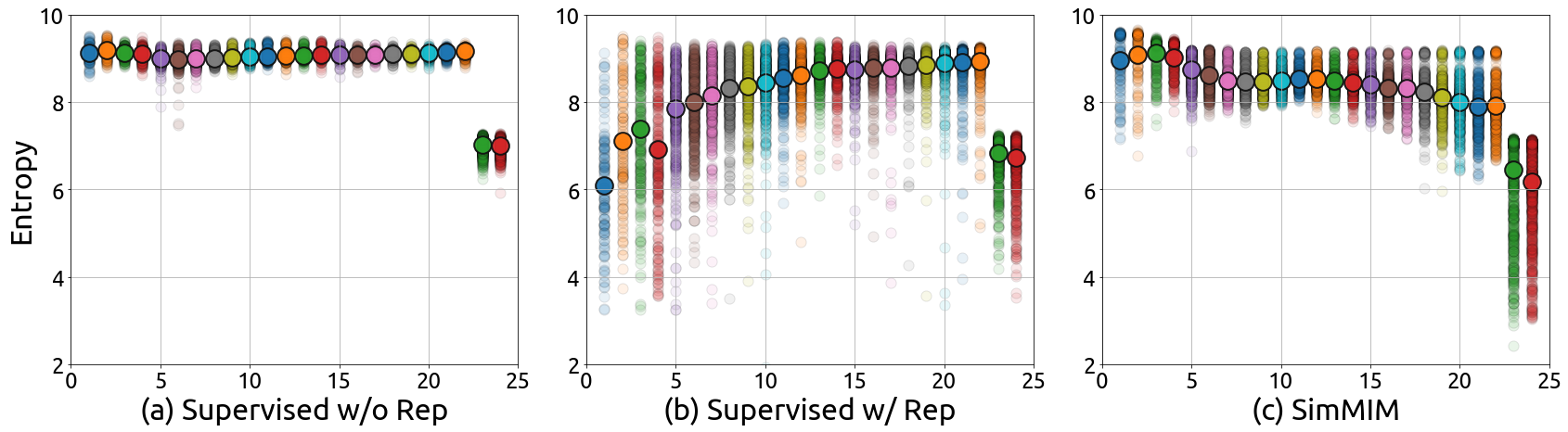}
    \caption{The entropy values in different channels (small dots) and the averaged entropy values (large dots) w.r.t the layer number on (a) supervised model without the re-parametrization trick, (b) supervised model with the re-parametrization trick, and (c) SimMIM model, with RepLKNet-31B as the backbone architecture.}
    \label{fig:replknet_entropy}
\end{figure}

\paragraph{Focused Kernels or Broad Kernels?} 
As shown in  Figure~\ref{fig:replknet_entropy}, with the re-parametrization trick, the supervised RepLKNet-31B model (b) has very focused attention in lower layers, but broader attention in higher layers. But for the MIM model (c), the entropy values in different kernels focus diversely in all layers, that some kernels are more focused and some kernels have very broad attention. These observations well match that in the Vision/Swin Transformers.

\begin{figure}[h]
    \centering
    \includegraphics[width=1.\linewidth]{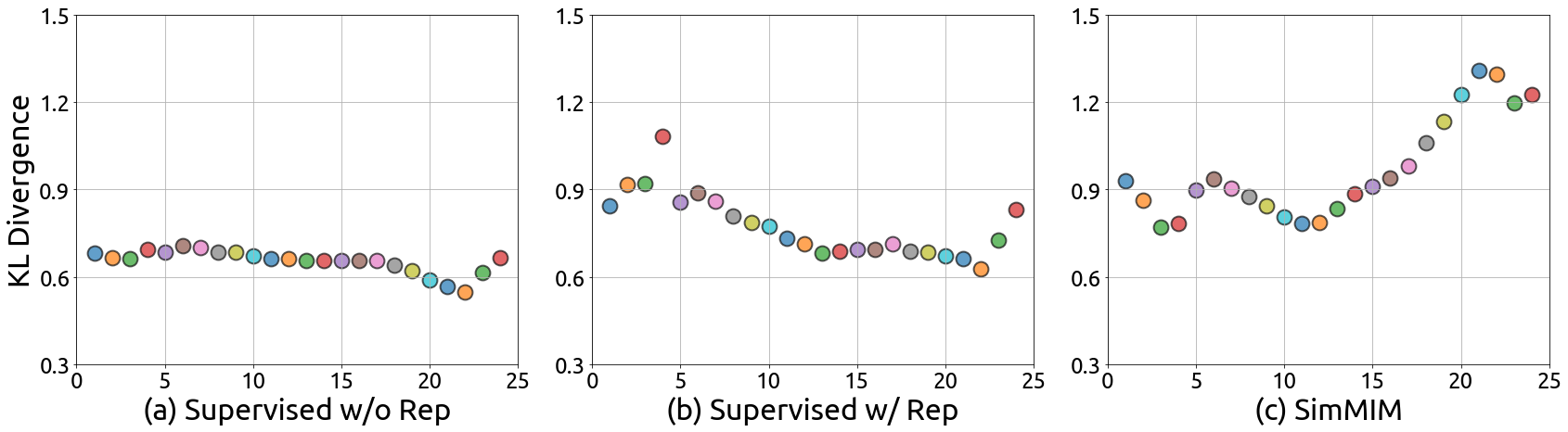}
    \caption{The averaged KL divergence in each layer w.r.t the layer number on (a) supervised model without the re-parametrization trick, (b) supervised model with the re-parametrization trick, and (c) SimMIM model, with RepLKNet-31B as the backbone architecture.}
    \label{fig:replknet_kl_divergence}
\end{figure}

\paragraph{Diversity across Different Kernels}

Interestingly, in Figure~\ref{fig:replknet_attention_distance}, it seems that the different kernels in both supervised model with the re-parametrization trick and SimMIM model have diverse averaged aggregated distance. But in Figure~\ref{fig:replknet_kl_divergence}, we could clearly observe that the diversity on different convolution kernels of SimMIM model (c) is remarkably larger than that of supervised counterparts (b), especially for the deeper layers.

\section{Detailed Results on Semantic Understanding Tasks}

\begin{table}[h]
  \centering
  \footnotesize
    \addtolength{\tabcolsep}{.5pt}
  \begin{tabular}{c|cccc|ccccccc}
    \toprule
    Methods &\rotatebox{75}{Food101} &\rotatebox{75}{Birdsnap} &\rotatebox{75}{Stanford Cars} &\rotatebox{75}{FGVC Aircraft} &\rotatebox{75}{Oxford Pets} &\rotatebox{75}{Caltech101} &\rotatebox{75}{Flowers102} &\rotatebox{75}{DTD} &\rotatebox{75}{SUN397} &\rotatebox{75}{CIFAR10} &\rotatebox{75}{CIFAR100} \\
    \midrule
    1K-SUP & 93.2 & 81.7& 88.6 & 83.0 & 95.9 & 91.9 & 97.7 & 80.3 & 72.3 & 99.1 & 91.0\\
    1K-MIM & 94.2 & 83.7 & 89.2 & 83.5 & 90.9 & 85.5 & 91.4 & 73.4 & 70.8 & 99.2 & 91.4\\
    \bottomrule
  \end{tabular}
  \vspace{0.3em}
  \caption{Detailed comparisons of MIM and supervised (SUP) pre-trained models on Kornblith 12-dataset classification benchmark~\cite{kornblith2019better} with SwinV2-B as the backbone. We follow~\cite{kornblith2019better} to report top-1 accuracy ($\uparrow$) and mean per-class accuracy ($\uparrow$) for specific datasets. Results on the multi-label dataset Pascal Voc 2007 are not included, whose evaluation metric is not compatible with others.}
  \label{tab-details-k12}
\end{table}

\renewcommand{\arraystretch}{1.15}
\begin{table}[h]
  \centering
  \addtolength{\tabcolsep}{-.3pt}
  \begin{tabular}{c|ccccc|c}
    \bottomrule
    \multirow{2}{*}{pre-train}  & \multicolumn{5}{c|}{Concept Generalization (CoG)} & 
    \multirow{2}{*}{iNat18}\\
    & $L_1$ & $L_2$ & $L_3$ & $L_4$ & $L_5$ & \\
    \hline
    RAND   & 79.4 & 76.7 & 73.1 & 72.7 & 68.5 & 76.5 \\
    1K-SUP & 79.4 & 76.2 & 72.7 & 72.5 & 68.4 & 77.7 \\
    1K-MIM & 79.6 & 77.1 & 73.6 & 73.0 & 69.1 & 79.6 \\
    \toprule
  \end{tabular}
  \caption{Detailed comparisons of randomly initialized model (RAND), MIM and supervised (SUP) pre-trained models on  Concept Generalization (CoG) benchmark~\cite{sariyildiz2021cog} and a fine-grained classification dataset iNaturalist-18~\cite{van2018inaturalist} with SwinV2-B as the backbone. Top-1 accuracy ($\uparrow$) is reported.
  }
  \label{tab-details-cog-inat}
  \vspace{-1.0em}
\end{table}

Detailed comparisons of Kornblith 12-dataset classification benchmark~\cite{kornblith2019better} and Concept Generalization (CoG) benchmark~\cite{sariyildiz2021cog} with a fine-grained classification dataset iNaturalist-18~\cite{van2018inaturalist} using SwinV2-B as the backbone, are shown in Table~\ref{tab-details-k12} and \ref{tab-details-cog-inat}, respectively. These results are already discussed in Section 4.1 of the main paper.

\section{Comparisons on Combined Task of Semantic Segmentation}

	\begin{table*}[t]
		\centering\setlength{\tabcolsep}{6pt}
		\footnotesize
		\begin{tabular}{cc|cc|cc}
			\bottomrule
			\multirow{3}*{backbone} & \multirow{3}*{pre-train} & \multicolumn{2}{c|}{Object Det. (COCO)} & \multicolumn{2}{c}{Semantic Seg. (ADE-20K)} \\
			 & & \multicolumn{2}{c|}{Mask R-CNN} & UperNet & Mask2former\\
			 & & AP$^{box}$ & AP$^{mask}$ & mIoU & mIoU\\
			\hline
			\multirow{2}*{SwinV2-B} & $1$K-SUP &$51.9$ & $45.7$ & $50.9$ & $52.3$ \\
			 & $1$K-MIM & $52.9$ & $46.7$ & $49.3~{\color{ForestGreen} (-1.6)}$ & $51.7~{\color{ForestGreen}(-0.6)}$ \\
			\toprule
		\end{tabular}
		\caption{Comparisons of MIM and supervised (SUP) pre-trained models on the combined tasks of object detection and semantic segmentation. We report the AP$^{box}$ ($\uparrow$) and AP$^{mask}$ ($\uparrow$) for the object detection and instance segmentation tasks, mIoU ($\uparrow$) for the semantic segmentation task.} 
		\label{tab:seg}
		\vspace{-0.8em}
	\end{table*}

We further select semantic segmentation on ADE-20K as another combine task which simultaneously performs both semantic understanding and geometric learning. For this task, we select two different frameworks, UperNet~\cite{xiao2018upernet} and Mask2former~\cite{mask2former} for evaluation. The detailed settings are shown in Section~\ref{sec:exp-detail}.

Results are shown in Table~\ref{tab:seg}. Different to COCO, we find that the supervised pre-trained model slightly outperforms the MIM counterpart on ADE-20K semantic segmentation. Therefore, for the combined tasks, it may be difficult to predict which pretrained model will perform better. But if the model gets larger, MIM models still have the unique advantage that MIM tasks are harder to be overfitted than supervised tasks~\cite{he2021masked,xie2021simmim}, which is beyond the scope of this paper. 
Also, we can observe that the performance gap between supervised and MIM models on Mask2former is smaller than that of UperNet ($-1.6$ v.s. $-0.6$). This may be due to that Mask2former decomposes the semantic segmentation task into object localization and recognition tasks, while MIM is better at object localization tasks, as shown in Figure 7 of the main paper.

\section{Detailed Settings}
\label{sec:exp-detail}

\begin{table}[h]\small
    \centering
    \begin{tabular}{c|cccccc}
    \toprule
    \multirow{2}{*}{\textbf{Hyperparameters}} 
    & \multicolumn{2}{c}{\textbf{RAND}}
    & \multicolumn{2}{c}{\textbf{1K-SUP}} & \multicolumn{2}{c}{\textbf{1K-MIM}}
    \\
    & \textbf{CoG (1-5)} & \textbf{iNat18} 
    & \textbf{CoG (1-5)} & \textbf{iNat18}
    & \textbf{CoG (1-5)} & \textbf{iNat18}
    \\
    \hline
    Input size & \multicolumn{6}{c}{224}
    \\
    Window size & \multicolumn{6}{c}{14}\\
    Patch size & \multicolumn{6}{c}{4}\\
    \hline
    Training epochs
    & 300 & 300 & 100 & 100& 100& 100
    \\
    Warm-up epochs & \multicolumn{6}{c}{20}\\
    Layer decay & 1.0 & 1.0 & 0.85 & 0.9 & 0.8 & 0.75 \\
    Batch size & \multicolumn{6}{c}{2048}     \\
    Optimizer & \multicolumn{6}{c}{AdamW}     \\
    Base learning rate & 2e-3 & 4e-3 & 2e-4 & 1.6e-3 & 5e-3 & 1.6e-2 \\
    Weight decay & 0.05 & 0.1 & 0.05 & 0.1 & 0.05 & 0.1 \\
    Adam $\epsilon$ & \multicolumn{6}{c}{1e-8}\\
    Adam $\beta$ & \multicolumn{6}{c}{(0.9, 0.999)}\\
    Learning rate scheduler & \multicolumn{6}{c}{Cosine}\\
    \hline
    Gradient clipping & \multicolumn{6}{c}{5.0}\\
    Stochastic depth & 0.5 & 0.5 & 0.2 & 0.2 & 0.2 & 0.2 \\
    Label smoothing & \multicolumn{6}{c}{0.1}\\
    \hline
    Rand crop scale & \multicolumn{6}{c}{(0.08, 1)} \\
    Rand resize ratio & \multicolumn{6}{c}{(3. / 4., 4. / 3.)} \\
    Rand horizontal flip & \multicolumn{6}{c}{0.5} \\
    Color jitter & \multicolumn{6}{c}{0.4}\\
    Rand augment & \multicolumn{6}{c}{9 / 0.5} \\
    Rand erasing prob. & \multicolumn{6}{c}{0.25}\\
    Mixup prob. & \multicolumn{6}{c}{0.8} \\
    Cutmix prob. & \multicolumn{6}{c}{1.0} \\
    \bottomrule
    \end{tabular}
    \vspace{0.5em}
    \caption{Detailed settings and hyperparameters for fine-tuning on CoG (1-5) and iNat18 with supervised and MIM pre-trained models.}
    \label{table:setting-finetune}
\end{table}

\noindent\textbf{Concept Generalization benchmark (CoG).}
The Concept Generalization benchmark (CoG) consists of five 1k-category datasets splitted from ImageNet-22K, which have increasing semantic gaps with ImageNet-1K, from $L_1$ to $L_5$.
On the CoG datasets, for a fair comparison, we first fine-tune the models on the CoG $L_1$ training set and search for the best hyper-parameter based on the validation top-1 accuracy of CoG $L_1$, and then directly apply the searched setting to CoG $L_2$ to $L_5$ and report the top-1 accuracy. 
The detailed hyperparameters are shown in Table~\ref{table:setting-finetune}.

\noindent\textbf{Kornlith et al's 12-dataset benchmark (K12) and iNaturalist-18 (iNat18).} 
On the K12 dataset, we follow the previous standard settings~\cite{kornblith2019better} to use training set and validation set to search for the best hyper-parameters, and then merge the training and validation sets as the final training set with the searched best hyper-parameters, and evaluate the final trained models on the test set. And we adopt standard splits of train/val/test sets as in~\cite{kornblith2019better}. For Aircraft, Pets, Caltech-101, Oxford 102 Flowers, the mean-per-class accuracy metric is adopted, for other datasets, the top-1 accuracy is adopted.
For K12, we follow~\cite{kornblith2019better} to select the optimal learning rate, weight decay, layer decay, and drop path rate. 
In pilot experiments, we find that for 1K-SUP pre-trained models, the drop path rate can be fixed as $0.2$, and for 1K-MIM pre-trained models, on smaller datasets like Stanford Cars, FGVC Aircraft, DTD,  Caltech101, Flowers102, and Oxford Pets, drop path rate is first fixed as $0.0$ and fixed as $0.2$ for other datasets. And the weight decay can be fixed as $0.05$.
Then we do a grid search on learning rate and layer decay. For 1K-MIM pre-trained models, our grid consists of 5 approximately logarithmically spaced learning rates between $1.25e$-$4$ and $2.5e$-$3$ and 3 equally spaced layer decay between $0.75$ and $0.95$. For 1K-SUP pre-trained models, our grid consists of 5 approximately logarithmically spaced learning rates between $2.5e$-$5$ and $5e$-$4$ and 3 equally spaced layer decay between $0.75$ and $0.95$. Then we adjust the learning rate, layer decay, and drop path rate in the neighborhood of the best setting in the grid search to get the final results. 

The iNat18 dataset includes 437,513 training images and 24,426 validation images, with more than 8,000 categories. 
The detailed hyperparameters of iNat18 are shown in Table~\ref{table:setting-finetune}.

\noindent\textbf{Pose estimation.}
    We compare the performance of MIM and supervised pre-trained models on the COCO~\cite{lin2014microsoft} and CrowdPose~\cite{li2019crowd} dataset. For the COCO dataset, We train the models on the $train2017$ set ($57K$ training images) and report the performance of the COCO $val2017$ split ($5K$ images), COCO $test$-$dev2017$ split ($20K$ images). For the CrowdPose dataset, following the DEKR~\cite{geng2021dekr}, we train the models on the CrowdPose train and val sets ($12K$ training images) and evaluate on the test split ($8K$ images). The standard average precision based on OKS is adopted as the evaluation metric for all datasets. 
	
	We adopt the heatmap-based top-down pipeline. We upsample the last feature of the backbone by deconvolutions and predict the heatmaps at 4$\times$ resolution like Simple Baseline~\cite{xiao2018sbaseline}. 
	
	In the ablation study on the number of the dropped layers in the section 3.1.3 of the main paper, we feed the feature at the different layers in the third stage of SwinV2-B into the pose head. We observe that when we use the feature at the ninth layer, the downstream performances of the supervised pre-trained model and MIM pre-trained model are almost comparable, so we use this model as the baseline of the experiments of randomly sampling pre-trained weights in the section 3.2 of the main paper. In the experiments of randomly sampling pre-trained weights, we randomly sample the weights of nine layers from the weights of the eighteen pre-trained layers in the third stage and then load them to the first nine layers. 
	
	The data augmentations include random flipping, half body transformation, random scale ($0.5$, $1.5$), random rotation ($-40^{\circ}$, $40^{\circ}$), grid dropout and color jitterring (h=$0.2$, s=$0.4$, c=$0.4$, b=$0.4$). The input image size is $256\times 256$ by default. We use the AdamW~\cite{Loshchilov2019adamw} optimizer with the base learning rate $5e$-$4$ and the weight decay $5e$-$2$. The learning rate is dropped to $5e$-$5$ at the $120th$ epoch. We totally train the models for $150$ epochs. We use a layer decay of $0.9$/$0.85$ for Swin-B/L and the DropPath~\cite{huang2016deep} of $0.3$/$0.5$ for Swin-B/L.
	The batch size is $512$.
	
    For the COCO dataset, we use the person detection results from the previous methods~\cite{sun2019hrnet, xiao2018sbaseline} for a fair comparison. For the CrowdPose dataset, we use a cascade mask-rcnn~\cite{cai2019cascadercnn} with Swin-B backbone trained on the COCO detection dataset to generate the person detection results. We use the UDP~\cite{huang2020udp} to reduce the quantization errors brought by the heatmaps and use flip testing by averaging the heatmaps predicted by the original and flipped images during the inference. 
	
	\noindent\textbf{Depth estimation.}
	We evaluate the performance of MIM and supervised pre-trained models on the NYUv2~\cite{nathan2012nyuv2} and KITTI~\cite{andreas2013kitti} monocular depth estimation datasets. 
	The NYUv2 dataset includes $464$ indoor scenes captured by a Microsoft Kinect camera. The official training split ($24K$ images) is used for training and we report the RMSE (Root Mean Square Error) on the $654$ testing images from $215$ indoor scenes.
	The KITTI dataset contains various driving scenes. The Eigen split~\cite{eigen2014depth} contains $23K$ training images and $697$ testing images.
	To compare with the previous approaches~\cite{ranftl2021dpt, kim2022glpdepth}, we set the maximum range as $10$m for NYUv2 and $80$m for KITTI.
	
	The head of the depth estimation is the same as the head of the pose estimation and is comprised of three deconvolutions (with BN and ReLU) and a normal convolution. The kernel and filter of the deconvolution are $2$ and $32$, respectively. 
	
     Similar to the GLPDepth~\cite{kim2022glpdepth}, we use the following data augmentations: random horizontal flip, random brightness (-$0.2$, $0.2$), random gamma (-$0.2$, $0.2$), random hue (-$20$, $20$), random saturation (-$30$, $30$), random value (-$20$, $20$) and random vertical CutDepth. We randomly crop the images to $480\times 480$ size for NYUv2 dataset and $352\times 352$ size for KITTI dataset. The optimizer, layer decay, and DropPath is the same as the pose estimation. The learning rate is scheduled via polynomial strategy with a factor of $0.9$. The minimal learning rate and the maximal learning rate are $3e$-$5$ and $5e$-$4$, respectively. The batch size is $24$. The total number of epochs is $25$. We use the flip testing and sliding window test for the SwinV2 backbone. We average the prediction of the two square windows for NYUv2 dataset and the sixteen square windows for KITTI dataset. 
	 
	\noindent\textbf{Video Object Tracking.}
	Following the previous arts, we train the models on the train splits of four datasets GOT10k~\cite{huang2021got10k}, TrackingNet~\cite{muller2018tracknet}, LaSOT~\cite{fan2019lasot}, and COCO~\cite{lin2014microsoft} and report the success score (SUC) for the TrackingNet dataset and LaSOT dataset. For the GOT10k test set, we report the average overlap as the evaluation metric. The GOT10k and the TrackingNet are two short-term large-scale benchmarks, the GOT10K test set contains $180$ video sequences, and the TrackingNet test set contains $511$ video sequences. The LaSOT is a long-term tracking benchmark and has $280$ video sequences with an average length of about $2500$ frames.
	
	We use the SwinTrack~\cite{lin2021swintrack} to evaluate our pre-trained models. The data augmentations and the training settings Strictly follow SwinTrack~\cite{lin2021swintrack}. 
	We sample $131072$ pairs per epoch and train the models for $300$ epochs. We use the AdamW optimizer with a learning rate of $5e$-$4$ for the head, a learning rate of $5e$-$5$ for the backbone, and a weight decay of $1e$-$4$. We decrease the learning rate by a ratio of $0.1$ at the $210$th epoch. We set the sizes of search images and templates as $224\times 224$ and $112\times 112$. The batch size is $160$. The inference process is the same as the SwinTrack~\cite{lin2021swintrack}.
	
	\noindent\textbf{Object Detection.} Following~\cite{xie2021simmim}, we adopt a Mask-RCNN~\cite{Mask-rcnn} framework to evaluate the pre-trained models on COCO object detection. All models are trained with a 3$\times$ schedule (36 epochs). We utilize an AdamW~\cite{kingma2014adam} optimizer with a learning rate of 6e-5 for supervised model and a learning rate of 8e-5 for MIM model, a weight decay of 0.05 and a batch size of 32 for both models. Following~\cite{simple_copy_paste,liu2021swin}, we employ a large jittering augmentation (1024 $\times$ 1024 resolution, scale range [0.1, 2.0]). The window size is set to 14 for both models and drop path rate is set to 0.3 for supervised model and 0.1 for MIM model. AP$^{box}$ and AP$^{mask}$ are reported for comparison.
	
	\noindent\textbf{Semantic Segmentation.} Following~\cite{liu2021swin}, an UPerNet~\cite{xiao2018upernet} framework is used for ADE-20K semantic segmentation. We use an AdamW~\cite{kingma2014adam} optimizer with a learning rate of 8e-5 for supervised model and a learning rate of 1e-4 for MIM model, a weight decay of 0.05 and a batch size of 32 for both models. Both models utilize a layer-wise learning rate decay of 0.95. All models are trained for 80K iterations with an input resolution of 640 $\times$ 640 and a window size of 20. The drop path rate is set to 0.3 for supervised model and 0.1 for MIM model.  In inference, a single-scale test using resolution of 2560 $\times$ 640 is employed.
	
	Besides, we also adopt Mask2Former~\cite{mask2former} to evaluate the pre-trained models on ADE-20K semantic segmentation. We use an AdamW~\cite{kingma2014adam} optimizer with a base learning rate of 1e-4 for supervised model and a base learning rate of 3e-4 for MIM model, a weight decay of 0.05 and a patch size of 16 for both supervised and MIM models. The learning rate of backbone is multiplied by a factor of 0.1. All models are trained for 160K iterations with an input resolution of 512 $\times$ 512, a scale ratio range from 0.5 to 2, a window size of 8, and a drop path rate of 0.3. In inference, the input resolution will be set to 2048 $\times$ 512. mIoU is reported for comparison for both UPerNet and Mask2Former.

\end{document}